\documentclass[10pt]{article} %
\usepackage[accepted]{tmlr}

\usepackage{amsmath,amsfonts,bm}

\def\eqref#1{equation~\ref{#1}}

\def\1{\bm{1}}

\def\vs{{\bm{s}}}

\DeclareMathAlphabet{\mathsfit}{\encodingdefault}{\sfdefault}{m}{sl}
\SetMathAlphabet{\mathsfit}{bold}{\encodingdefault}{\sfdefault}{bx}{n}

\def\sR{{\mathbb{R}}}

\newcommand{\E}{\mathbb{E}}

\DeclareMathOperator*{\argmax}{arg\,max}

\newcommand{\normal}{\mathcal{N}}

\def\shownotes{1}  %
\ifnum\shownotes=1
\newcommand{\authnote}[2]{ [\emph{#1: #2}]}
\else
\newcommand{\authnote}[2]{}
\fi
\ifnum\shownotes=1

\else

\fi

\newcommand{\update}[1]{#1}%

\def\oursolution{MERMAIDE}%

\def\agentalgo{f}
\def\hiddenstateclass{H}
\def\latentfunc{g}

\newcommand{\eq}[1]{\begin{align}#1\end{align}}

\newcommand{\fr}[2]{\frac{#1}{#2}}
\newcommand{\brck}[1]{\left(#1\right)}

\newcommand{\brcksq}[1]{\left[#1\right]}
\newcommand{\brckcur}[1]{\left\{#1\right\}}

\def\Planner{Principal}
\def\capsPlanner{Principal}
\def\planner{principal}

\def\agent{agent}
\def\Bernoulli{\textrm{Bern}}
\def\uniform{\mathcal{U}}

\def\iagent{i}
\def\iplanner{p}
\def\itime{t}

\def\action{a}

\def\actionset{A}

\def\cost{c}
\def\costcoeff{\alpha}
\def\df{\gamma}
\def\epochiter{e}
\def\hidden{h}
\def\horizon{T}
\def\hist{\mathcal{H}}

\def\intervention{\reward'}
\def\interventionpolicy{\policy^\iplanner}

\def\mean{\mu}
\def\meanadj{\tilde{\mu}}
\def\numagents{n}
\def\nummetatrainepochs{E_{\textrm{train}}}

\def\nummetatraintrainsteps{K_\textrm{train}}

\def\optstep{k}
\def\numoptsteps{K}
\def\numtasks{N}
\def\obj{J}
\def\obs{o}
\def\policy{\pi}

\def\reward{r}

\def\realreward{\tilde{r}}

\def\state{s}
\def\statevec{\vs}
\def\stateset{S}

\def\suboptgap{\delta}
\def\task{\tau}
\def\taskset{\mathcal{T}}

\def\testtaskset{\taskset_{\textrm{test}}}
\def\testsetsize{\numagents_{\textrm{test}}}
\def\testK{\numoptsteps}
\def\traintaskset{\taskset_{\textrm{train}}}
\def\trainsetsize{\numagents_{\textrm{train}}}
\def\transition{\mathcal{P}}
\def\trans{\transition}

\def\latentpar{h}
\def\ppolicypar{\theta}
\def\apolicypar{\omega}
\def\worldmodel{\hat{\policy}_\apolicypar}

\def\task{\tau}

\def\dataset{\mathcal{D}}

\def\agentlearningalgo{f}
\usepackage{caption}
\usepackage{subcaption}

\usepackage{microtype}
\usepackage{graphicx}
\usepackage{booktabs} %

\usepackage{hyperref}

\usepackage{adjustbox}

\usepackage{amsmath}
\usepackage{amssymb}
\usepackage{mathtools}
\usepackage{amsthm}
\usepackage{algorithm}

\let\classAND\AND
\let\AND\relax
\usepackage{algorithmic}

\let\AND\classAND
\AtBeginEnvironment{algorithmic}{\let\AND\algoAND}

\usepackage[capitalize,noabbrev]{cleveref}

\usepackage{comment}
\usepackage{float}

\usepackage{blkarray}
\newcounter{assumptioncntr}

\newcounter{remarkcntr}

\usepackage{url}

\title{\oursolution{}: Learning to Align Learners using Model-Based Meta-Learning}

\author{\name Arundhati Banerjee\thanks{Work done during an internship at Salesforce} \email arundhat@cs.cmu.edu \\
    \addr School of Computer Science\\
    Carnegie Mellon University
    \AND 
    \name Soham Phade\thanks{Work done while at Salesforce} \email soham\_phade@berkeley.edu \\
    \addr Wayve Technologies Ltd
    \AND 
    \name Stefano Ermon \email ermon@cs.stanford.edu \\
    \addr Department of Computer Science\\
    Stanford University
    \AND 
    \name Stephan Zheng \email stephan@asari.ai\\
    \addr Asari AI
}

\def\month{11}%
\def\year{2023}%
\def\openreview{\url{https://openreview.net/forum?id=H5VRvCXCzf}}%

\begin{document}

\maketitle

\begin{abstract}
We study how a \planner{} can efficiently and effectively intervene on the rewards of a previously unseen \emph{learning} agent in order to induce desirable outcomes.
This is relevant to many real-world settings like auctions or taxation, where the \planner{} may not know the learning behavior nor the rewards of real people.
Moreover, the \planner{} should be few-shot adaptable and minimize the number of interventions, because interventions are often costly.
We introduce \oursolution{}, a model-based meta-learning framework to train a \planner{} that can quickly adapt to out-of-distribution agents with different learning strategies and reward functions.
We validate this approach step-by-step. 
First, in a Stackelberg setting with a best-response agent, we show that meta-learning enables quick convergence to the theoretically known Stackelberg equilibrium at test time, although noisy observations severely increase the sample complexity.
We then show that our model-based meta-learning approach is cost-effective in intervening on bandit agents with unseen explore-exploit strategies. 
Finally, we outperform baselines that use either meta-learning or agent behavior modeling, in both $0$-shot and $1$-shot settings with partial agent information.
\end{abstract}

\section{Introduction}
In many application domains, such as revenue maximization in auctions \citep{milgrom2004putting}, economic policy design for social welfare \citep{doi:10.1126/sciadv.abk2607} or optimizing skill acquisition in personalized education \citep{maghsudi2021personalized}, a \planner{} seeks to incentivize an \emph{adaptive} agent to achieve the \planner{}'s goal.
In this work, we assume that both the \planner{} and the agent are learners and the \planner{} incentivizes by directly intervening on the rewards of the agent.
However, the \planner{} does not know neither the exact value of the agent's rewards nor its learning algorithm (or its parameters).
For instance, a government (\planner{}) may want to incentivize the use of environmentally-friendly products by levying green taxes, but needs to consider that people (agents) may change their consumption behavior as taxes change.
Here, common agent models based on rationality or forms of bounded rationality often do not fully describe real-world behavior. 
Hence, interacting with the agents is required for the \planner{} to \emph{learn} how agents change their behavior, but such interactions are not ``free'' or without risk.
For example, measuring the impact of a tax change on consumers takes effort, while it may incur economic costs if the new tax is unfair or not well calibrated. 
\update{
In a similar vein, for designing intelligent tutoring systems, the tutor (\planner{}) has to adapt to different students (\agent{}s) and learn to incentivize the students to understand different concepts, without the designed curriculum being too difficult or too easy (incurring cost for the \planner{}). 
In personalized health-monitoring apps, the app (\planner{}) has to adapt to the user's (\agent{}) lifestyle and preferences to maybe create a plan that can incentivize the user to fulfill their health-related goals. 
}

To reduce the need for real-world interactions, we can use simulations with deep reinforcement learning (RL) agents.
This is an attractive solution framework:
deep neural networks are expressive enough to imitate real-world entities and simulations can be run safely and at will.
Moreover, we can use deep RL to learn intervention policies that are effective even in the face of complex agent behaviors in \emph{sequential} \planner{}-agent problems. 

However, this approach also faces several challenges.
When deploying the learned policies in the real world, interventions can typically only be applied a few times, due to implementation costs,
and rarely under identical circumstances;
in contrast to simulations, we cannot reset the real world.
Even though \planner{}s may adapt their policies to new conditions, they cannot realistically know the true rewards or learning strategy of the agent.
Hence, our goal is to learn the \planner{}'s intervention policy that %
1) can perform well even when agents learn, 
2) can be quickly adapted,
3) is robust to distribution shifts in agent behaviors, and 
4) is effective despite having only \emph{partial information}.

\paragraph{Contributions.}
To address these challenges, we propose \oursolution{} (\textbf{Me}ta-lea\textbf{r}ning for \textbf{M}odel-based \textbf{A}daptive \textbf{I}ncentive \textbf{De}sign), a deep RL approach %
that 1) learns a world model %
and 2) uses gradient-based meta-learning to learn a \planner{} policy that can be quickly adapted to perform well on unseen test agents.
We consider a \emph{\planner{} and single-agent} setting %
wherein the \planner{} intervenes \emph{at a cost} on the agent's learning process to incentivize the agent to learn to act to achieve the \planner{}'s objective.
We assume that the \agent{} behaves in a first-order strategic manner and the \planner{} in a second-order strategic manner. %
Here, the agents optimize their experienced rewards and minimize their regret, but do not account for their influence on the \planner{}'s actions.
In contrast, the \planner{} intervenes explicitly as to influence the \agent{}'s actions.

\paragraph{Claims.}
In summary, our work demonstrates the challenges faced in the \planner{} and single \agent{} setup with adaptive learning \agent{}s. 
\oursolution{} is an effective framework to address these challenges.

Our empirical results validate the advantages, but also show the limitations of a purely model-based approach to address the non-stationarity in the environment for the \planner{} and the adaptive \agent{}s. 
Our proposed solution, \oursolution{}, shows that model-based meta-learning is able to further improve on a purely model-based approach and learns a cost-effective few-shot adaptable intervention policy for the \planner{}.

More specifically, in this work, we empirically verify the following claims about \oursolution{}:
\paragraph{Claim 1.} 
In a single-round Stackelberg game between the \planner{} and an \agent{}, 
\update{
where the \planner{} (leader) acts first by deciding whether to intervene or not, and the \agent{} (follower) acts second according to a best-response policy \citep{von2010market}, our
}
meta-trained \planner{} reliably finds solutions that one-shot adapt well with a best response \agent{} under both perfect and noisy observations for the \agent{} and the \planner{}. 
Furthermore, the \planner{}'s out-of-distribution performance depends on its observable information about the agent. 
\paragraph{Claim 2.} 
In the multi-armed bandit setting, \oursolution{} finds well-performing reward intervention policies in a repeated interactive setup with the adaptive bandit agents. 
\oursolution{}'s test-time performance and robustness against out-of-distribution bandit learners depends on the \agent{}'s level of exploration and their level of pessimism in the face of uncertainty, which are unknown to the \planner{}. 
In particular, this holds for both $\testK{} = 0$-shot and $\testK{} = 1$-shot evaluation. 

To the best of our knowledge, prior works studying the \planner{}-\agent{} setup, even in an adaptive setting, overlook several of these problems with more restrictive assumptions on either the observations available or the types of \agent{}s and \planner{}.

\section{Related Work}
\paragraph{Mechanism design.}
The \planner{}-\agent{} problem \citep{eisenhardt1989agency} studies how a \planner{} can incentivize and align the \agent{} with the \planner{}'s goals, in particular in the face of information asymmetry (e.g., the \planner{} does not know the \agent{}'s rewards).
This relates to the general mechanism design setting in which a \planner{} typically interacts with multiple agents \citep{hurwicz2006designing}. 
Learning a mechanism with \agent{}s who also learn is a bilevel optimization problem, which is NP-hard \citep{ben1990computational, sinha2017review}. 
Possible solution techniques include branch-and-bound and trust regions \citep{colson2007overview}.
In particular, solving bilevel optimization using joint learning of the mechanism and the \agent{}s can be unstable, since \agent{}s continuously adapt their behavior to changes in the mechanism. 
This can be stabilized using curriculum learning  \citep{zheng_ai_2020}, but generally bilevel problems remain challenging, especially with nonlinear objectives or constraints. 
Moreover, unlike with typical curriculum learning \citep{bengio2009curriculum}, in our setting, the \planner{}’s intervention essentially changes the task for the adaptive agent, thus presenting a non-stationary learning problem for both the \planner{} and the \agent{} at each time step.

\paragraph{Adaptive mechanism design.}
Previous work in mechanism design usually does not consider learning how to learn to incentivize across agents of different types.
\citet{10.1145/1151454.1151480} found that a form of meta-learning that adapts the learning process itself can design English auctions (sequential bidding) that perform better with adaptive bidders who are loss-averse, and is still effective when the distribution of bidder behaviors (slowly) shifts.
Prior work has studied algorithms for incentivizing exploration in bandit agents \citep{chen2018incentivizing, wang2021incentivizing} where the principal can assign incentives to temporary and myopic agents for choosing different arms so that the principal can passively determine the global preferences of the agent population. 
\citet{shi2021almost} extend this line of work to consider non-myopic strategic agents, but unlike our principal-agent problem formulation, they assume that the agent is aware of the principal's incentive before choosing an arm and the agent always selects the incentivized arm. 
Moreover, their setting does not focus on few-shot generalization to unseen test agents. %
Our work expands on this theme by explicitly modeling \agent{}s that learn, considering shifts in the \emph{learning algorithm} of the \agent{}s, and using deep RL with meta-learning. 
This combination enables learning incentives that generalize well across more complex tasks.

\paragraph{Meta-learning and inverse RL.}%
In recent years, gradient-based meta-learning has proven effective in learning initializations for complex policy models that generalize well to unseen tasks \citep{finn2017model, nagabandi2018learning}.
\citet{luketina2022meta} study %
meta-gradients for adapting in environments with controlled sources of non-stationarity, but ignore non-stationarity from interactions between strategic agents that learn. %
Prior works in imitation learning \citep{argall2009survey} and inverse RL \citep{abbeel2004apprenticeship} assume access to expert demonstrations with a fixed policy to imitate or learn the reward function of, whereas \citet{jacq2019learning, ramponi2020inverse} consider inverse RL with observers from learners that improve their policies, but they do not feature a \planner{} that actively intervenes. 
In contrast, our \planner{} aims to learn a policy that can strategically \emph{alter} the behavior of such demonstrators (our agents), who are themselves \emph{learning} during an episode of the demonstration. 
Recently, \citet{NEURIPS2020_171ae1bb} studied meta-learning for bandit policies, while \citet{https://doi.org/10.48550/arxiv.2106.14866} %
introduced the inverse bandit setup for learning from low-regret demonstrators.
However, these works do not consider shifts in the bandit learning algorithm %
between training and test time.

\paragraph{Modeling agents.}
A key challenge in multi-agent learning is that each \agent{} experiences a non-stationary environment if other \agent{}s are learning.
As such, \agent{}s can benefit from having a \emph{world model}, e.g., to know what the policy or value function of the other agents are. 
World models can stabilize multi-agent RL \citep{lowe_multi-agent_2017} and enable higher-order learning methods \citep{foerster_learning_2018}, and are a form of model-based RL.
However, this may require a large amount of observational data or prior knowledge, which may be hard to acquire.
We show that world models make \planner{}s much more efficient in our setting.

\section{Problem formulation}
\paragraph{Overview.}
We model a \emph{\planner{}} who aims to incentivize an \emph{\agent{}} to (learn to) execute the \planner{}'s preferred action.
To do so, the \planner{} can \emph{intervene} and change the \agent{}'s rewards at a cost. 
Without interventions, the \agent{} may learn to prefer an action different than the \planner{}'s.
 
For example, consider consumers who can use either environmentally ``clean'' or ``dirty'' goods. 
Indifferent at first, consumers may gradually learn to prefer dirty goods if those are consistently cheaper than clean ones, whereas the government may want them to prefer clean goods.
Here, the \agent{}'s reward is the negative of the cost of consumption, for instance, and an intervention changes the price of goods through taxes or subsidies.
If we can use a simulation, the \planner{} can compute an optimal intervention. 
However, the simulation might be inaccurate and real-world agents might behave differently. 
As an example of such test-time distribution shift, simulated \agent{}s may be quick to change their consumption patterns, while real \agent{}s may be slow.
A ``good'' \planner{} (trained in a simulation) could quickly be fine-tuned to intervene more in the latter case and adapt quickly if such behavior is observed during deployment.

In particular, we focus on \emph{learning} a \planner{} policy that %
can adapt quickly at test-time 
(e.g., deploying taxes and subsidies in the real world), and that is effective when the \agent{}'s learning algorithm differs from that %
during training.%

We now formalize this setting.
In this work, we focus on \agent{}s in a stateless environment for ease of exposition.
For all variables and their meaning, see \cref{sec:appendix_notation}. %

\paragraph{The agent.} 
The \agent{}s are characterized by their action space $\actionset$ and a base reward function $\reward : \actionset \rightarrow \sR$. 
We call it base reward as the \agent{} experiences an \emph{intervened} reward 
\eq{
\realreward_\itime\brck{\action_\itime} = \reward\brck{\action_\itime} + \intervention_\itime\brck{\action_\itime},
}
where the intervention $\intervention_\itime$ is provided externally (by the \planner{}) for the agent action $\action_\itime$. 
We index time as $\itime = 1, \ldots, \horizon$. 
At each time step $\itime$, the \agent{}'s policy $\policy_\itime$
computes a distribution over its actions based on the observations for the agent up to timestep $\itime$ and executes $\action_\itime\sim\policy_\itime$.
We assume that the \planner{} has a \emph{preferred action} $\action^*\in\actionset$ that the agent should execute, whereas the \agent{}'s optimal policy can prefer a different action than $\action^*$ without intervention. 
Finally, at time $\itime$, %
the \agent{} learns using an update rule $\agentlearningalgo: \brck{\policy_\itime, \action_\itime, \realreward_\itime} \mapsto \policy_{\itime + 1}$ 
to maximize the \agent{}'s intervened 
rewards, e.g., under UCB \citep{lai1985asymptotically}, $\agentlearningalgo$ updates the confidence bounds for the action selected at time $\itime$.

\paragraph{The \planner{}.}
In this work, from the \planner{}'s point of view, the \emph{world (environment)} consists of the \agent{} who maximizes $\realreward$.
A standard assumption is that \agent{}s are rational and they may have a private state (referred to as its \emph{type}) which the \planner{} cannot see.
Although the agent faces a stateless problem, \emph{the \planner{} faces a stateful problem with partial observability}. 
The full state $\state\in\stateset$ includes the \planner{}'s internal state $\hidden^\iplanner_\itime$ (e.g., the \planner{}'s belief about the value of the private \agent{} information), and all information about the \agent{}, including its past actions, reward function, and policy model; often, the latter two are private.

More formally, the \planner{} can be modeled as a POMDP $\brck{\stateset, \obs^\iplanner, \actionset^\iplanner, \reward^\iplanner, \df, \trans}$. 
It receives observations $\obs^\iplanner$ (a part of the world state $\state$), $\actionset^\iplanner$ is its action space of interventions, $\reward^\iplanner$ is its reward, $\df$ is a discounting factor, and $\trans$ are the environment dynamics, e.g., as caused by the \agent{}'s actions.
At time $\itime$, the \planner{} samples an action $\bm\action^\iplanner_\itime \sim \policy^\iplanner\brck{\bm\action^\iplanner_\itime|\obs^\iplanner_{\itime-1}, \hidden^\iplanner_{\itime-1}}$ which determines its intervention on each possible agent action $\action$, i.e. $\bm\action^\iplanner_\itime = \brcksq{\intervention_1,\dots,\intervention_{|\actionset|}}$. 

\paragraph{Adaptive intervention policy learning}%
To model distribution shift at test time, we follow the meta-learning terminology \citep{finn_model-agnostic_2017} and view each distinct \agent{} as a \emph{task} $\task^\iagent$.
The \planner{} has access to a \emph{train set of \agent{}s} $\task^\iagent\in\traintaskset; \iagent = 1, \ldots, \trainsetsize$ and is evaluated on a \emph{test set of \agent{}s} $\task^\iagent \in \testtaskset; \iagent = 1, \ldots, \testsetsize$.
\emph{We emphasize that during a task, both the \planner{} and \agent{} may learn and adapt, both at train and test time.}

Here, we focus on two key challenges: $\testK$-shot adaptation and distribution shift. 
First, the \planner{} gets only $\testK$ episodes for fine-tuning for each test task (but can train indefinitely for each train task).
Second, the \planner{} faces two types of distribution shift: 1) across tasks and 2) intra-task non-stationarity. 
The train and test tasks may differ (significantly) in their temporal distribution of actions, e.g., due to different \agent{} updates $\agentlearningalgo$ or the \agent{} rewards $\reward_\itime$ being centered around different values (e.g., average price levels are higher in the real world vs in the simulation).
Within a task, the \agent{}'s learning is affected by the \planner{}'s interventions that change its reward $\realreward$.
This gives rise to non-stationarity in the \agent{}'s environment, as its learning objective may shift over time. 
These forms of distribution shift distinguish our %
adaptive intervention policy learning
setting from most prior work in meta-learning, which often assume stationarity within a task and also assume similar task distributions at train and test times.

\paragraph{Objectives.}
The \planner{}'s objective is to maximize how often test-time \agent{}s choose $\action^*$ during learning and have them converge to a policy that always chooses $\action^*$.
To do so, the \planner{} aims to maximize the cost-adjusted test-time return %
$\obj_\textrm{test}^\iplanner\brck{\policy^\iplanner, \policy^{\iagent}} = \sum_{\itime=1}^\horizon \df^{\itime-1} (\reward^\iplanner_{\itime} - \costcoeff\cost_\itime)$, 
where the \agent{} executes its (optimal) policy $\policy^{\iagent}\brcksq{\interventionpolicy}$ in response to $\interventionpolicy$ 
and the \planner{} incurs a cost $\cost_\itime$ if it intervenes. 
$\reward^\iplanner_\itime = \bm{1}\brcksq{\action_\itime = \action^*}$,  $\costcoeff > 0$.

\begin{align}\label{eq:planner-objective}
{\interventionpolicy}^* &= \argmax_{\interventionpolicy} \E_{\task^\iagent\in\testtaskset}\E_{\interventionpolicy}\E_{\policy^{\iagent}\brcksq{\interventionpolicy}}\brcksq{ 
\sum_{\itime=1}^\horizon \df^{\itime-1} (\reward^\iplanner_{\itime} - \costcoeff\cost_\itime)
} %
\end{align}

A simple cost function is
$
\cost_\itime = \bm{1}\brcksq{\intervention_\itime \neq 0}$, 
i.e., the cost is constant across non-trivial interventions, where $\costcoeff > 0$ is a constant.
Note that if intervention were free ($\cost_t = 0$), a trivial solution is to always add a large $\intervention\brck{\action^*} \gg 0$ for its preferred action $\action^*$, such that it always yields the highest reward. 
Hence, we focus on learning non-trivial strategies when intervention is costly, which forces the \planner{} to strategically alter the \agent{}'s learning behavior.

During an episode of $\horizon$ time steps, each agent $\iagent$ starts with a uniformly initialized action probability distribution $\policy_0^\iagent$ and optimizes $\policy_\itime^\iagent$ subject to interventions $\interventionpolicy$ to maximize its return:
$\E_{\policy^\iagent}\E_{\policy^\iplanner}
\brcksq{ \left. \sum_{\itime=1}^\horizon  \realreward^\iagent_{\itime}\brck{\action^\iagent_\itime,\action^\iplanner_\itime} \right.}$.
Here, we assume that $\horizon$ and $\df$ are sufficiently large so the \agent{} converges to its optimal policy under $\realreward$, using its learning algorithm $\agentlearningalgo$. 
That is, we assume that the objective in \cref{eq:planner-objective} is sufficient to describe the \planner{}'s objective of ensuring the \agent{} converges to preferring $\action^*$ at some $\itime < \horizon$.

In the $\testK$-shot adaptation setting, at test time, the \planner{} gets $\testK$ episodes to interact with any \agent{}, each episode of length $\horizon$ steps.
The \planner{} has a fixed policy during an episode and it can update its policy at the end of an episode. 
The \agent{} is reset across episodes, and within each episode, the agent follows its own learning strategy in response to the \planner{}'s interventions.
On the $\testK+1^{\text{th}}$ episode, the \planner{} evaluates its $\testK$-shot adapted policy on the \agent{}.
Note this assumes that the \planner{} has a separate copy of the test time \agent{} for evaluation.

\section{\oursolution{}: Learning to Align Learners}

\label{sec:our_algo}

\begin{figure*}[t]
    \centering
    \begin{subfigure}{0.5\linewidth}
        \centering
        \includegraphics[scale=0.18]{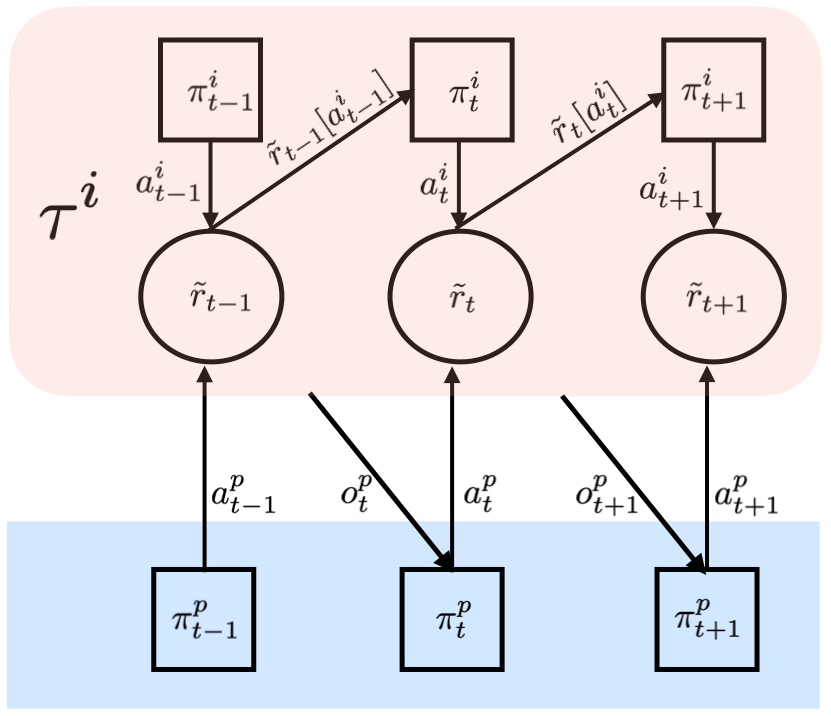}\;\;%
        \label{fig:modelgraph}
    \end{subfigure}%
    \begin{subfigure}{0.5\linewidth}
        \centering
        \includegraphics[scale=0.18]{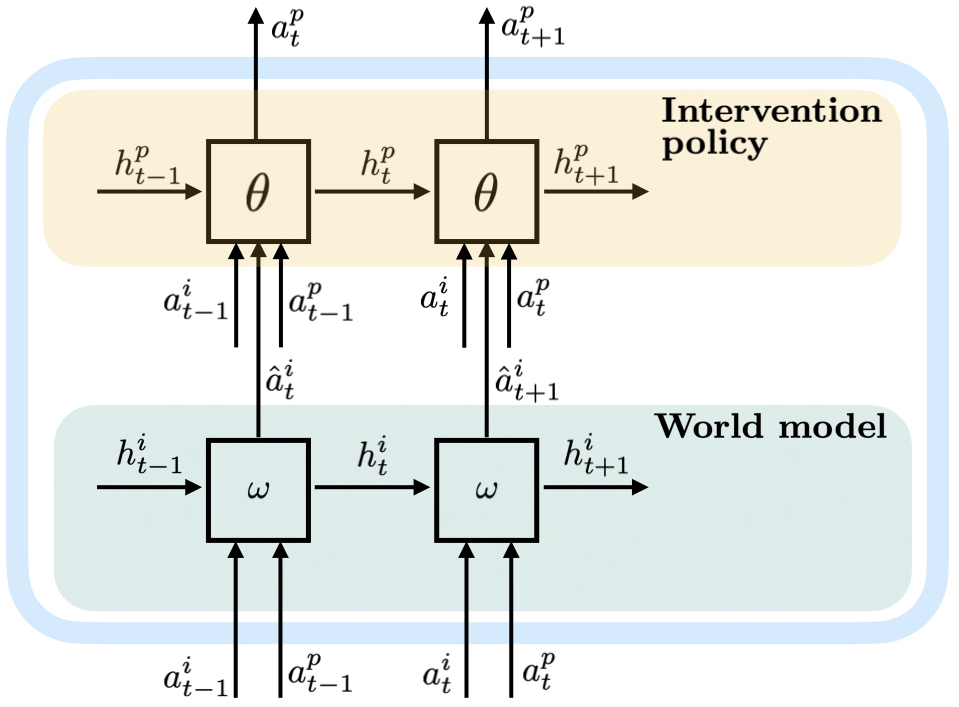}
        \label{fig:modeloutline}
    \end{subfigure}
    \caption{Overview of \textbf{\oursolution{}.} Left: Flow of \planner{} and agent observables, rewards, and actions. Right: The \planner{}'s world model and intervention policy. 
    Also see \cref{algo:our_algo}.}
    \label{fig:ourmodel}
\end{figure*}

\begin{algorithm}[t]
\caption{
    \oursolution{} (Notations also in \cref{tab:proof_notation})
}
\begin{algorithmic}[1]
\small
\STATE{Initialize \planner{} ($\ppolicypar_0$, $\apolicypar_0$), and hidden states $\hidden^\iagent_0, \hidden^\iplanner_0$. }
\FOR{meta-train epoch $\epochiter = 1,\dots,\nummetatrainepochs$}
\STATE{Update world model parameters $\apolicypar = \apolicypar_\epochiter$ (\cref{eq:wm_train}).}%
\FOR{agents (tasks) $\iagent = 1,\dots,\trainsetsize$}
\STATE{Initialize \agent{}: ($\mean^\iagent,\policy^\iagent_0$), task specific \planner{} policy parameter  $\ppolicypar\brck{\task^\iagent_0}=\ppolicypar_\epochiter$.
}
\FOR{$\optstep = 1,\dots,\nummetatraintrainsteps$} %
\FOR{time \itime{} = $1,\dots,\horizon$} 
\STATE{Predict $\hat{\action}^\iagent_{\itime} = \argmax_{\action^\iagent_\itime} \worldmodel\brck{\action^\iagent_\itime | \action^\iagent_{\itime-1},\action^\iplanner_{\itime-1},\hidden^\iagent_{\itime-1}}$}%
\STATE{Intervention: 
$\meanadj^\iagent = \mean^\iagent + \action^\iplanner_\itime, \quad
\action^\iplanner_\itime \sim \policy^\iplanner_{\ppolicypar\brck{\task^\iagent_\optstep}}\brck{ \action^\iplanner_\itime | \action^\iagent_{\itime-1}, \action^\iplanner_{\itime-1}, \hat{\action}^\iagent_\itime,\hidden^\iplanner_{\itime-1}}$.
}
\STATE{Agent acts: $\action^\iagent_\itime\sim\policy^\iagent_\itime$ and receives reward $\reward^\iagent_\itime \sim \normal\brck{\meanadj^\iagent, \sigma^2}$. $\policy^\iagent_{\itime} \mapsto \policy^\iagent_{\itime+1}$.} 
\ENDFOR
\STATE{Locally update $\ppolicypar\brck{\task^\iagent_\optstep} \mapsto \ppolicypar\brck{\task^\iagent_{\optstep+1}}$. \COMMENT{Using REINFORCE.} }\label{line:local_update_our_algo}%
\ENDFOR 
\COMMENT{Rollout for meta-update; $\dataset_{\text{meta}}\brck{\task^\iagent}=\{\}$}
\FOR{$\itime = 1,\dots,\horizon$}
\STATE{Predict $\hat{\action}^\iagent_{\itime} = \argmax_{\action^\iagent_\itime} \worldmodel\brck{\action^\iagent_\itime | \action^\iagent_{\itime-1},\action^\iplanner_{\itime-1},\hidden^\iagent_{\itime-1}}$}%
\STATE{Intervention: $\meanadj^\iagent = \mean^\iagent + \action^\iplanner_\itime, \quad
\action^\iplanner_\itime \sim \policy^\iplanner_{\ppolicypar\brck{\task^\iagent_{\nummetatraintrainsteps}}}\brck{ \action^\iplanner_\itime | \action^\iagent_{\itime-1}, \action^\iplanner_{\itime-1}, \hat{\action}^\iagent_{\itime}, \hidden^\iplanner_{\itime-1}}$.
}
\STATE{Agent acts: $\action^\iagent_\itime\sim\policy^\iagent_\itime$, receives reward $\reward^\iagent_\itime \sim \normal\brck{\meanadj^\iagent, \sigma^2}$. Updates $\policy^\iagent_{\itime} \mapsto \policy^\iagent_{\itime+1}$.} 
\STATE{Collect 
$\dataset_{\text{meta}}\brck{\task^\iagent}\cup\brckcur{
\action^\iagent_{\itime},
\action^\iplanner_{\itime},
\policy^\iplanner_{\ppolicypar\brck{\task^\iagent_{\nummetatraintrainsteps}}}
}$
} 
\ENDFOR
\ENDFOR
\STATE{Meta-update $\ppolicypar_\epochiter \mapsto \ppolicypar_{\epochiter+1}$ using $\dataset_{\textrm{meta}}=\cup_{\task^\iagent}\dataset_{\textrm{meta}}\brck{\task^\iagent}$. 
\COMMENT{Using MAML.}
} \label{line:meta_update_ours}
\ENDFOR
\end{algorithmic}
\label{algo:our_algo}
\end{algorithm}

\oursolution{} learns an intervention policy to align the \agent{}'s preferred action with the \planner{}'s one, using:

1) a recurrent \emph{world model} parameterized by $\apolicypar$ that outputs a distribution over an \agent{} $\iagent$'s actions at the next time step $\itime$: $\worldmodel\brck{\action^\iagent_\itime | \action^\iagent_{\itime-1},\action^\iplanner_{\itime-1},\hidden^\iagent_{\itime-1}}$, %
conditioned on the planner's intervention and the observed agent action at $\itime-1$. 
$\hidden^\iagent_{\itime-1}$ is the hidden world model state.
$\action^\iagent_\itime\sim\policy^\iagent_\itime$.

2) a recurrent \emph{intervention policy} which outputs a distribution over interventions $\action^\iplanner_{\itime} \sim \policy^\iplanner_{\ppolicypar}\brck{\action^\iplanner_\itime |\action^\iagent_{\itime-1}, \action^\iplanner_{\itime-1}, \hat{\action}^\iagent_{\itime}, \hidden^\iplanner_{\itime-1}}$, conditioned on its previous intervention, the observed agent action and the %
    world model's %
    predicted next agent action $\hat{\action}^\iagent_\itime = \max_{\action} \worldmodel\brck{\action | \action^\iagent_{\itime-1},\action^\iplanner_{\itime-1},\hidden^\iagent_{\itime-1}}$. $\hidden^\iplanner_{\itime-1}$ is the hidden state of the policy network.

We train this with gradient-based meta-learning and RL (\cref{algo:our_algo}). 
\update{
$\mean^\iagent$ indicates the mean or base reward function for agent $\iagent$ and $\tilde{\mean}^\iagent$ is the reward function after the principal's intervention $\action^\iplanner_\itime$ at time $\itime$. 
Please refer to \cref{tab:notation} and \cref{tab:proof_notation} for a comprehensive list of the notations used. 
}
Here, the \planner{} maximizes $\obj_\textrm{train}^\iplanner$ similar to the objective in \cref{eq:planner-objective}.
The base RL algorithm is REINFORCE \citep{williams1992simple} and the meta-learning update uses MAML \citep{finn_model-agnostic_2017}. %
The \agent{} optimizes its cumulative intervened reward (see \cref{sec:bandit-setting}). 
The world model $\worldmodel$ trains by maximizing the log-likelihood of the observed $\action^\iagent_\itime$, using Adam \citep{kingma2014adam}:
\begin{align}
    &\argmax_\apolicypar \E_{\action^\iplanner} \E_{\action^\iagent} \brcksq{
        \sum_{\itime = 1}^\horizon \log\worldmodel\brck{\action^\iagent_\itime | \action^\iagent_{\itime-1}, \action^\iplanner_{\itime-1}, \hidden_{\itime-1}^\iagent} 
    }\label{eq:wm_train}
\end{align}

Note that the \planner{}'s parameters $\ppolicypar$ are updated after each $\horizon$-step episode, while the \agent{} continuously learns during each episode.
Also, the \agent{} is reset in between episodes.
At time 0, the world model makes a prediction based on zero initialization.%
We use a single world model for all agents.
At meta-test time, only the intervention policy is updated by one-shot adaptation to a new agent (\cref{algo:K-ADAPT}).%

\section{Evaluating Meta-Learning for the \Planner{}} %
\label{sec:single-shot-setting}

We now compare the learning behavior of a \planner{} that is meta-trained (MAML) \citep{finn_model-agnostic_2017} versus one trained with standard policy gradients (RL), in a simple Stackelberg game between a \planner{} and an \agent{}. %
\update{
The Stackelberg game considers a leader-follower approach where the \planner{} (leader) acts first, deciding to intervene or not on the \agent{} (follower) who acts second according to a best-response policy. 
The key characteristic of a Stackelberg equilibrium is that once the leader has made their move, the followers have no incentive to deviate from their optimal responses, given the leader's action. 
This results in a stable, strategic outcome where each player is maximizing their utility or payoff based on the actions of their opponent. 
Our goal is to learn an intervention policy for the \planner{} that can adapt to different \agent{} types and find the Stackelberg equilibrium.
}

The \agent{}'s actions are ``cooperate'' and ``defect'', while the \planner{} can choose whether or not to intervene. 
Assuming the row player is the \agent{} and the column player is the \planner{}, we define a $2\times2$ payoff matrix 
\eq{
\begin{blockarray}{lcc}
& \text{No intervention (NI)} & \text{Intervene (IN)} \\
\begin{block}{l(cc)}
\text{Cooperate (C)} & u, 1 & u+1, 1-c \\
\text{Defect (D)} & 1-u, 0 & -u, -c \\
\end{block}
\end{blockarray}, \nonumber
}
where $u \in (0, 1)$ and $c$ is the cost of intervention ($c < 1$).
The \agent{}'s base payoff $u$ is its type. 
The \planner{} prefers cooperation:
it gets 1 if the \agent{} cooperates and 0 if the \agent{} defects (minus the cost $c$ if it intervenes).
Note that intervention incentivizes cooperation ($u+1>-u$).

We now analyze three scenarios with increasing difficulty:

    1) First, we assume that the \planner{} knows $u$.
    Here, there is a unique Stackelberg equilibrium at (C, NI) when $u\geq \fr{1}{2}$, and at (C, IN) when $u < \fr{1}{2}$. 
    
    2) Second, the \planner{} observes a noisy version of $u$. 
    In both cases, the \agent{} first observes the \planner{}'s action and plays its best-response (knowing the payoffs). 
    
    Note that both 1) and 2) are single-round Stackelberg settings. 
    
    3) Finally, we study a repeated (multi-round) %
    Stackelberg setting where the \agent{} cannot see the \planner{}'s action.
    Instead, we assume that the \agent{} keeps a running average for the experienced payoffs for each of its actions. 
    In an equivalent single-round setting this would correspond to the \planner{} committing to a mixed action and then the \agent{} choosing its best response.
    When $u\geq \fr{1}{2}$, the Stackelberg equilibrium occurs at (C, NI). %
    When $u<\fr{1}{2}$, at the Stackelberg equilibrium the \planner{} has a mixed action where it chooses to intervene for $\fr{2u+1}{2}$ fraction of times and the \agent{} always cooperates. 

\update{
Given these expected Stackelberg equilibrium strategies, our goal is to learn a policy for the \planner{} that predicts its probability of intervention and study the adaptivity of MAML vs RL trained policies on unseen \agent{}s. 
}
We set $c = 0.75$. %
Given a set of training \agent{}s with different types $u \sim \uniform\brck{0,1}$, the \planner{} learns the optimal policy parameters $\ppolicypar^* = \argmax_{\ppolicypar}\E_u\E_{\action^\iplanner\sim\policy^\iplanner_\theta\brck{u}}\brcksq{\reward^\iplanner\brck{\action^\iplanner}}$. %
Note that here we do not use a world model, rather we focus on model-free policy learning. 
We study the quality of the initialization $\ppolicypar_{\text{MAML}}$ vs $\ppolicypar_{\text{RL}}$ by evaluating the one-shot adaptation performance of the trained \planner{} on unseen test agents.

\begin{figure*}[t]
    \centering
    \begin{subfigure}{0.33\linewidth}
    \centering
        \includegraphics[width=\linewidth]{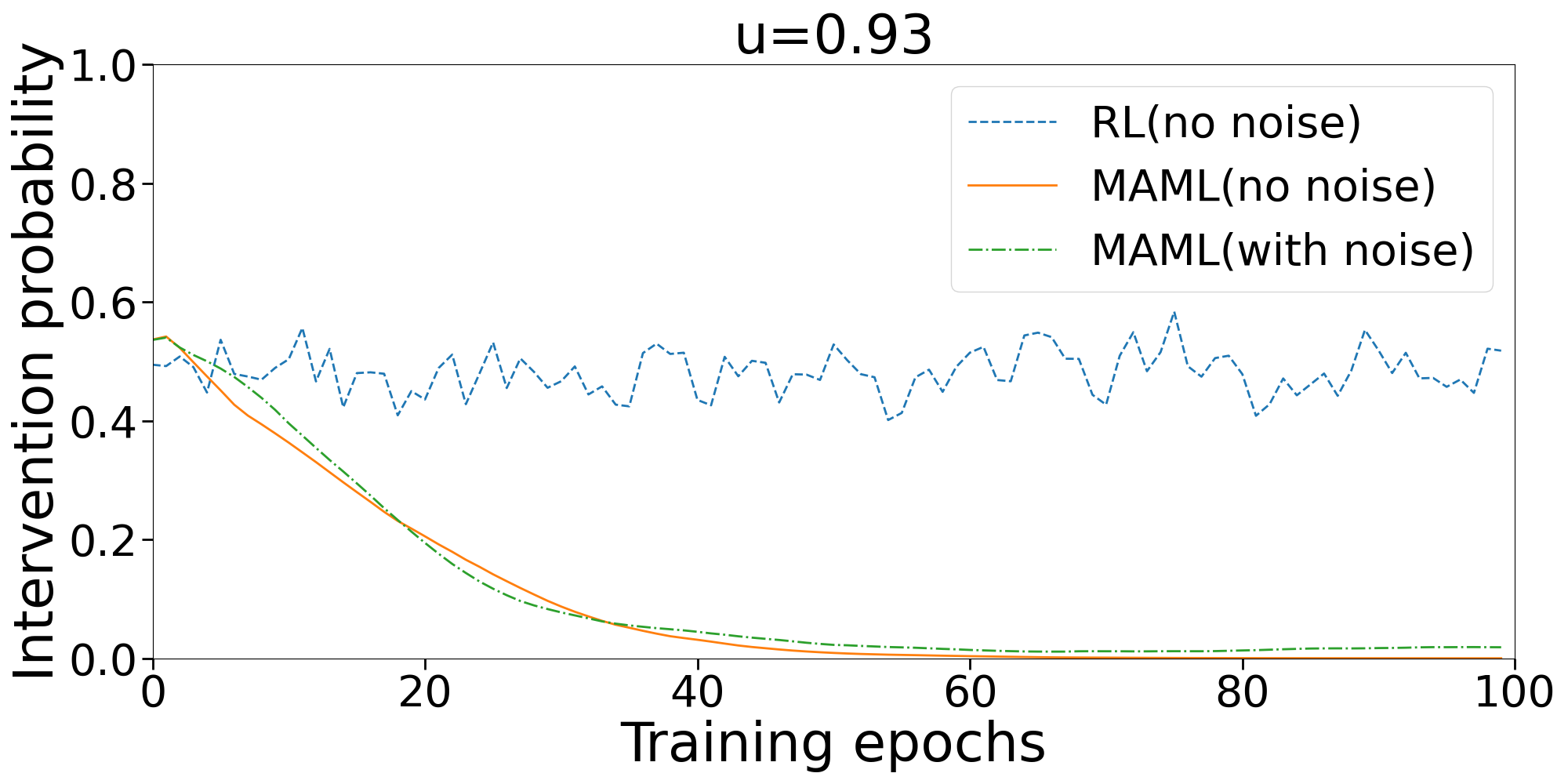}
        \caption{}%
        \label{fig:2x2_u_0.93}
    \end{subfigure}%
    \begin{subfigure}{0.33\linewidth}
    \centering
        \includegraphics[width=\linewidth]{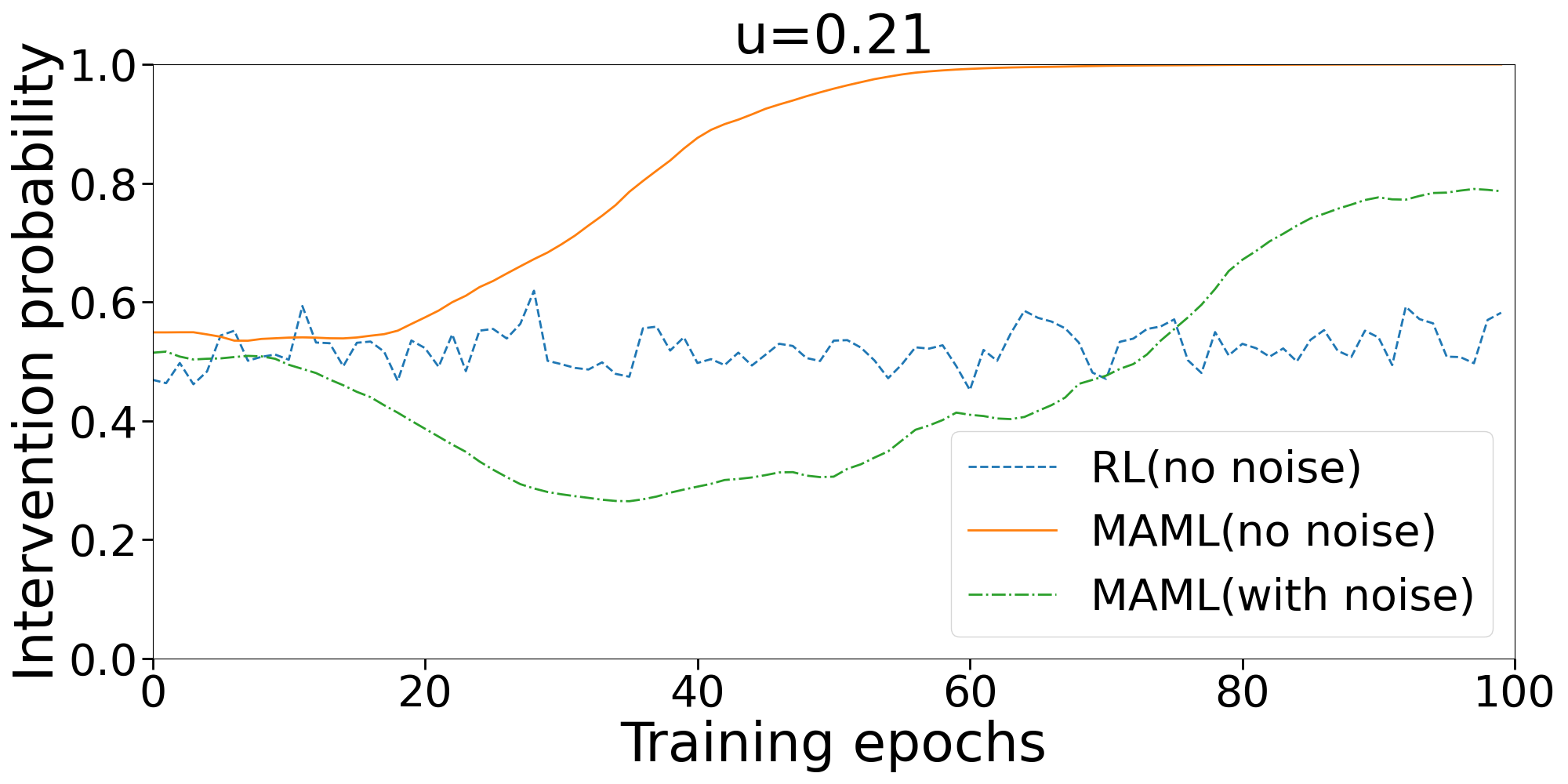}
        \caption{}%
        \label{fig:2x2_u_0.21}
    \end{subfigure}%
    \begin{subfigure}{0.33\linewidth}
    \centering
        \includegraphics[width=\linewidth]{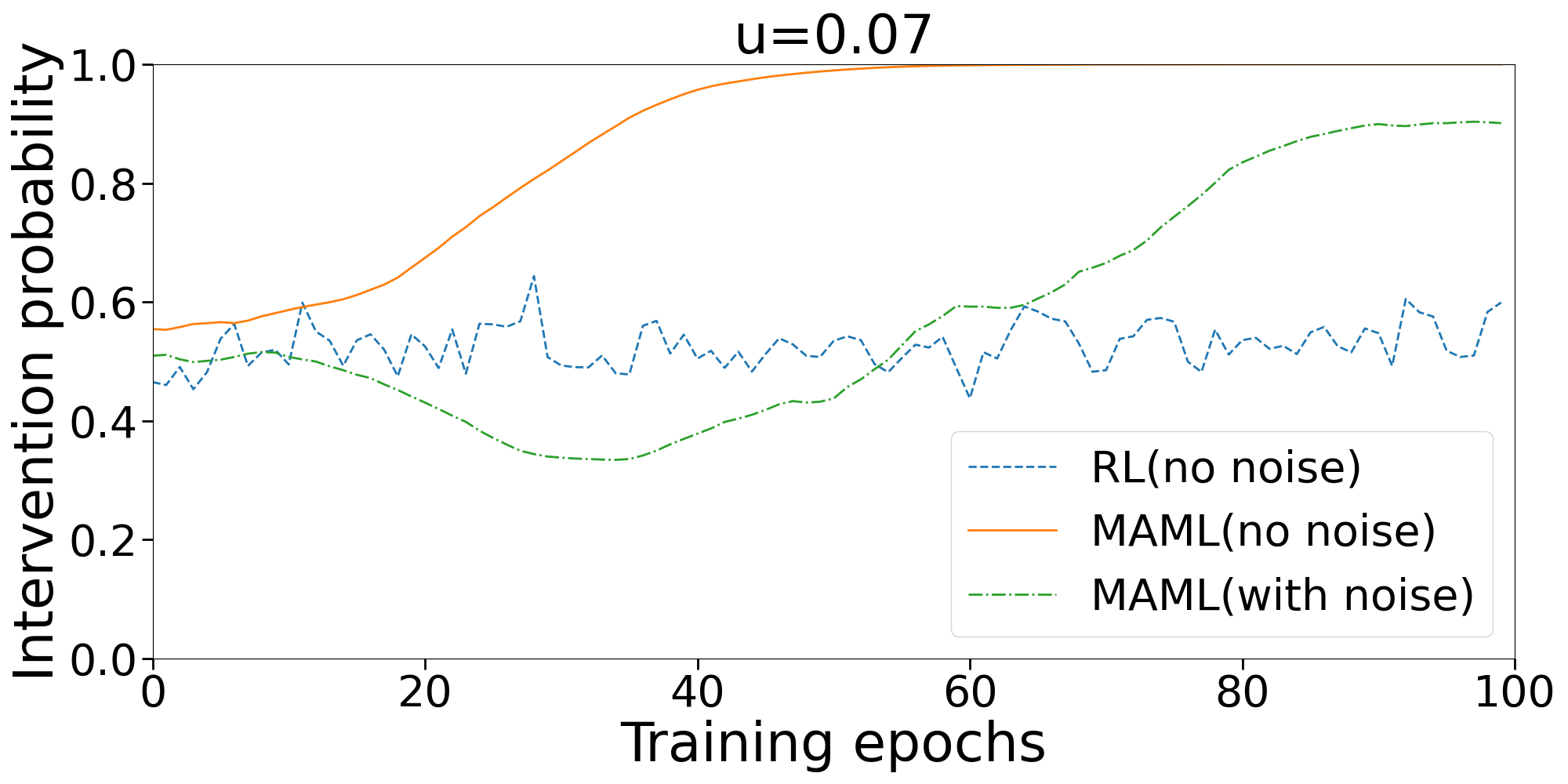}
        \caption{}%
        \label{fig:2x2_u_0.07}
    \end{subfigure}
    \caption{\textbf{Single round game.} REINFORCE (RL) does not adapt to expected Stackelberg equilibrium during evaluation. MAML's adaptability suffers under observation noise.} %
    \label{fig:2x2_viz_planner}
\end{figure*}

\paragraph{With perfect observability.}
In this setting, we assume that the \planner{} observes an \agent{}'s exact payoff parameter $u$.
\cref{fig:2x2_viz_planner} shows the \planner{}'s meta-test time probability of intervening with 3 different \agent{}s from the test set, across training epochs.
The \planner{} and \agent{} should be at different Stackelberg equilibria depending on the type $u$, as discussed above. 
We see that a \planner{} trained from scratch on the test \agent{}s using standard policy gradients is unable to adapt to different \agent{}s in a single-shot adaptation setting. 
In contrast, with meta-learning, the \planner{} learns a policy $\policy^\iplanner_{\theta_{\text{MAML}}}$ that is one-shot adaptable to \agent{}s of different types and converges to the correct Stackelberg equilibrium at meta-test time.

\begin{figure*}[t]
    \centering
    \begin{subfigure}{0.33\linewidth}
        \centering
       \includegraphics[scale=0.25]{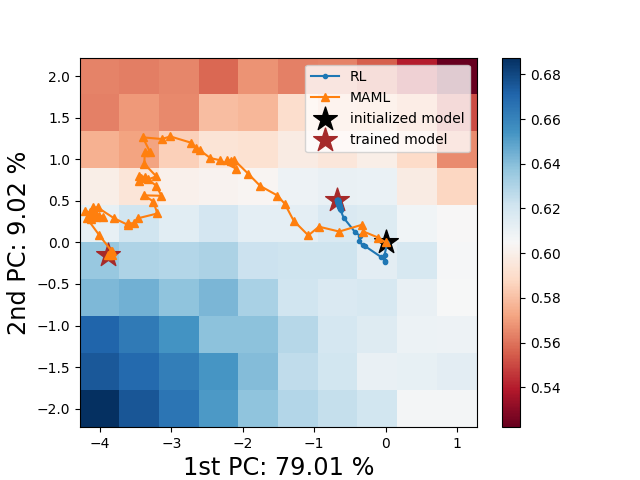}
        \caption{}
        \label{fig:opt_traj_set3}
    \end{subfigure}%
    \begin{subfigure}{0.33\linewidth}
    \centering
       \includegraphics[width=\linewidth]{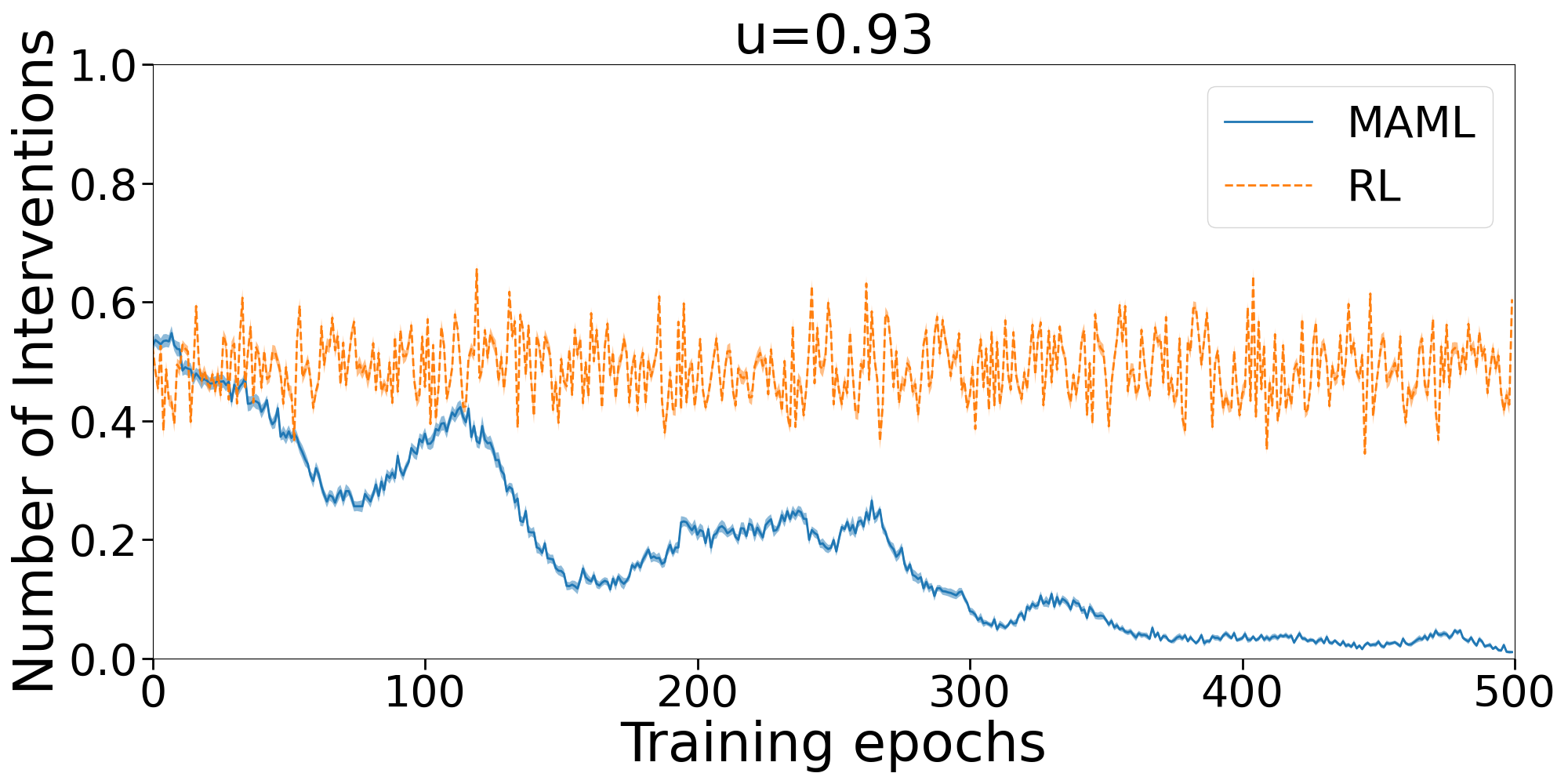}
        \caption{}
        \label{fig:2x2_u_0.93_set3}
    \end{subfigure}%
    \begin{subfigure}{0.33\linewidth}
    \centering
       \includegraphics[width=\linewidth]{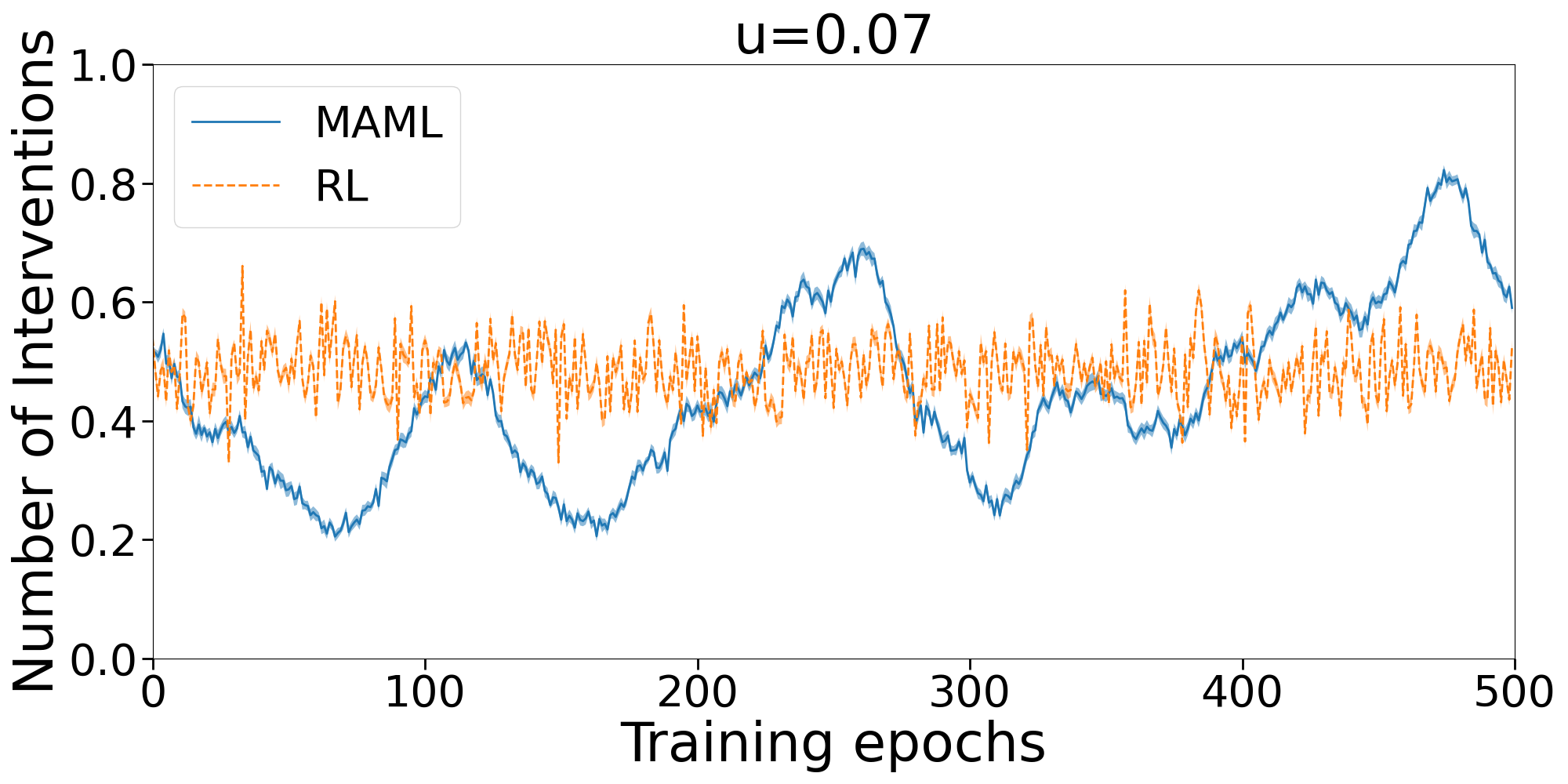}
        \caption{}
        \label{fig:2x2_u_0.07_set3}
    \end{subfigure}
    \caption{\textbf{Multi-round game.} (a) \capsPlanner{}'s optimization trajectory in the expected payoff landscape during training. Axes are PCA directions in the policy parameter space. \update{Colorbar indicates the \planner{}'s expected payoff over the training \agent{}s.} (b)(c) MAML adapts (single shot) to Stackelberg equilibrium with a best response agent. \update{The number of interventions are normalized by the episode length $\horizon=100$.}} 
    \label{fig:set3}
\end{figure*}

\paragraph{With noisy observations for the \planner{}.} 
Here, we emulate a \planner{} with partial observability of the \agent{}, by letting the \planner{} observe $u$ with added i.i.d. Gaussian noise. 
The \agent{} can see all payoffs and chooses the best response to achieve a Stackelberg equilibrium. 
\cref{fig:2x2_viz_planner} shows that with noisy observations, the meta-learned \planner{} policy requires more training time to be one-shot adaptable to the equilibrium intervention policy. 
This empirically indicates the increased difficulty of learning an adaptive intervention policy due to incomplete information about the agent, especially under limited adaptation time with unseen \agent{}s.
It therefore motivates us to adopt a \emph{model-based} approach for the \planner{} to better estimate the \agent{} type and learn an adaptive intervention policy.

Comparing \cref{fig:2x2_u_0.21} and \cref{fig:2x2_u_0.07}, %
when $u$ $<$ $\frac{1}{2}$, the difference in unintervened payoffs between the \planner{}'s preferred action ($u$) and the \agent{}'s preferred action ($1-u$) also impacts the one-shot adaptability of the \planner{} with noisy observations.
This informs our analysis in \cref{sec:bandit-setting}.

\paragraph{Multi-round repeated game with noisy rewards.}%
In this setting, the \planner{} and \agent{} repeatedly play an iterated game over $T=100$ steps.
In each round, the \planner{} observes the \agent{}'s type $u$ with added i.i.d. Gaussian noise.
The \agent{} cannot see the \planner{}'s actions, and best responds given its payoff estimates.
Whenever the \agent{} selects an action, it receives a noisy observation of the true payoff and updates its estimate.
Compared to the single-round setting, here the \agent{}'s best response behavior may change across rounds in the game depending on its observed payoffs, giving rise to non-stationarity in the \planner{}'s environment.
The planner, in turn, has to learn to intervene so that the \agent{}'s best response is to cooperate.

\cref{fig:opt_traj_set3} compares the optimization trajectory followed using 1) policy gradients and 2) meta-learning for the \planner{}.
Starting from the same initialization, the meta-learned policy's parameters lie in a region of the payoff landscape with a higher expected value over the training agents. 
Moreover, \cref{fig:2x2_u_0.93_set3} and \cref{fig:2x2_u_0.07_set3} show the one-shot adaptability of the \planner{}'s policy for two different \agent{} types at meta-test time.
Meta-learning helps learn a better intervention strategy that is robust to the \planner{}'s observation noise as well as the \agent{}'s evolving best response strategy.

This analysis shows the benefits of meta-learning over standard policy gradients in learning a \planner{}'s intervention policy at equilibrium with strategic agents and under partial observability for the \planner{} in a simple Stackelberg setting. 
Our observations motivate using meta-training and model learning in general \planner{}-agent settings. 
\oursolution{} uses both these components; next, we analyze its performance against learning agents with complex adaptive behavior under repeated interventions.

\section{Evaluating \oursolution{} on Bandits}
\label{sec:bandit-setting}

\begin{table*}[t]
    \centering
    \caption{
    \textbf{Test-time \planner{} performance on agents with different test-time learning parameters (3 random seeds).}
    Left column: \Planner{}'s algorithm (e.g., \oursolution{}), training \agent{} type (e.g., UCB with $\beta=0.17$).
    Other columns: Test-time scores (mean and standard error (s.e.)) on \agent{}s with the same algorithm, but different hyperparameters.
    \oursolution{} ($K=0$) indicates zero-shot evaluation.
    Rest are evaluated with one-shot adaptation ($K=1$).
    \oursolution{} outperforms all baselines in almost all cases in both the $\testK=0$ and $\testK=1$ cases, with $\testK = 1$ \planner{}s doing better than $\testK=0$.
    }
    \label{tab:results_table}
    \begin{tabular}{lrrrrr}%
    \textbf{Train on UCB, $\beta=0.17$} & Test on $\beta=0.17$ & $\beta=0.27$ & $\beta=0.42$ & $\beta=0.5$ & $\beta=0.67$ \\
    \emph{No intervention} & 3 (0) & 5 (0) & 8 (0) & 10 (0) & 12 (0)\\
    MF-RL & 119 (2) & 109 (2) & 98 (2) & 90 (2) & 77 (1)\\
    MF-MAML & 133 (2) & 125 (3) & 107 (1) & 97 (1) & 77 (0)\\
    WM-RL & 123 (7) & 112 (6) & 100 (4) & 92 (2) & 75 (1) \\
    {\oursolution{} ($K=1$)} & \textbf{154 (2)} & \textbf{151 (1)} & \textbf{129 (1)} & \textbf{129 (1)} & \textbf{87 (0)}\\
    {\oursolution{} ($K=0$)} & 148 (2) & 138(1) & 120(1) & 103(2) & 89(1) \\
    \midrule
    \textbf{Train on $\epsilon$-greedy, $\epsilon=0.1$} & $\epsilon=0.1$ & $\epsilon=0.2$ & $\epsilon=0.3$ & $\epsilon=0.4$ & $\epsilon=0.5$\\
    \emph{No intervention} & 3 (0) & 4 (1) & 7 (0) & 9 (1) & 11 (0) \\
    MF-RL & 115 (5) & 94 (4) & 54 (19) & 39 (6) & 22 (9)\\
    MF-MAML & 122 (4) & 97 (3) & 58 (5) & 40 (2) & 12 (1)\\
    WM-RL & 115 (4) & 94 (5) & 70 (1) & 55 (3) & \textbf{38 (1)}\\
    {\oursolution{} ($K=1$)} & \textbf{138 (1)} & \textbf{112 (2)} & \textbf{85 (3)} & \textbf{66 (2)} & {37 (4)}\\
    {\oursolution{} ($K=0$)} & 133 (2) & 109 (3) & 86 (2) & 65 (3) & 37 (1) \\
    \midrule
    \textbf{Train on UCB, $\beta=0.67$} & $\beta=0.17$ & $\beta=0.27$ & $\beta=0.42$ & $\beta=0.5$ & $\beta=0.67$\\
    \emph{No intervention} & 3 (0) & 5 (0) & 8 (0) & 10 (0) & 12 (0)\\
    MF-RL & 103 (3) & 101 (3) & 92 (2) & 85 (1) & 74 (1)\\
    MF-MAML & 124 (2) & 116 (1) & 102 (1) & 94 (1) & 80 (1) \\
    WM-RL & 100 (4) & 89 (0) & 85 (1) & 85 (1) & 74 (0)\\
    {\oursolution{} ($K=1$)} & \textbf{132 (1)} & \textbf{130 (1)} & \textbf{123 (2)} & \textbf{115 (2)} & \textbf{99 (1)}\\
    {\oursolution{} ($K=0$)} & 115 (3) & 114 (3) & 103 (4) & 100 (3) & 89 (3) \\
    \midrule
    \textbf{Train on $\epsilon$-greedy, $\epsilon=0.5$} & $\epsilon=0.1$ & $\epsilon=0.2$ & $\epsilon=0.3$ & $\epsilon=0.4$ & $\epsilon=0.5$\\
    \emph{No intervention} & 3 (0) & 4 (1) & 7 (0) & 9 (1) & 11 (0) \\
    MF-RL & 4 (5) & 2 (3) & 5 (0) & 11 (5) & 7 (1)\\
    MF-MAML & 2 (0) & 4 (0) & 6 (0) & 8 (1) & 11 (1)\\
    WM-RL & 102 (6) & 79 (10) & 68 (3) & 47 (1) & 30 (2)\\
    {\oursolution{} ($K=1$)} & 87 (42) & \textbf{102 (3)} & \textbf{78 (6)} & \textbf{69 (1)} & \textbf{46 (2)} \\
    {\oursolution{} ($K=0$)} & 113 (20) & 85 (15) & 71 (16) & 48 (14) & 21 (15) \\
    \end{tabular}
\end{table*}

\begin{table*}[t]
    \centering
    \caption{
    \textbf{$\bm\testK$$=$$0$-shot evaluation across different agent algorithms: test-time \planner{} scores (3 random seeds).}
    Left column: \planner{}'s algorithm (e.g., \oursolution{}), training \agent{} type (e.g., UCB, $\beta=0.42$).
    Other columns: Test-time scores (mean and s.e.) on \agent{}s with different algorithm and hyperparameters. 
    }
    \label{tab:zero_shot_cross_table}
    \begin{tabular}{lrrrrr}%
    \textbf{Train on UCB, $\beta=0.42$} & Test on $\epsilon=0.1$ & $\epsilon=0.2$ & $\epsilon=0.3$ & $\epsilon=0.4$ & $\epsilon=0.5$\\
    \emph{No intervention} & 3 (0) & 4 (1) & 7 (0) & 9 (1) & 11 (0) \\
    WM-RL & 91 (4) & 62 (8) & \textbf{68 (1)} & \textbf{28 (4)} &  -\\
    {\oursolution{} (ours)} & \textbf{103 (1)} & \textbf{67 (2)} & 30 (2) & 8 (1) & -\\
    \midrule
    \textbf{Train on $\epsilon$-greedy, $\epsilon=0.3$} & Test on $\beta=0.17$ & $\beta=0.27$ & $\beta=0.42$ & $\beta=0.5$ & $\beta=0.67$ \\
    \emph{No intervention} & 3 (0) & 5 (0) & 8 (0) & 10 (0) & 12 (0)\\
    WM-RL & 127 (7) & 95 (2) & 80 (5) & 80 (5) & 61 (4)\\
    {\oursolution{} (ours)} & \textbf{138 (2)} & \textbf{102 (6)} & \textbf{116 (2)} & \textbf{96 (5)} & \textbf{77 (2)} \\
    \end{tabular}
\end{table*}

We now consider a \planner{} intervening sequentially on an adaptive no-regret learner \agent{}, modeled by an $|\actionset|$-armed bandit instance with action set $\actionset$ having base reward $\bm\reward = \brcksq{\reward_1, \dots, \reward_{|\actionset|}}$.
At each time step $\itime$, the \agent{} chooses an arm $\action$ and gets a reward sampled from $\normal\brck{\reward_\action, \sigma^2}$.
We assume $\reward_\action\in\brck{0,1}\;\forall\action$. 
The \agent{} aims to maximize its cumulative reward over a horizon of $\horizon$ steps.
The \agent{} can only observe the reward for the chosen action, and hence faces a explore-exploit dilemma addressed by bandit algorithms like UCB \citep{lai1985asymptotically}. 
We assume there is a unique arm $\tilde{\action}$ with the highest base reward: $\tilde{\action} = \argmax_\action \reward_\action$, i.e., the \agent{}'s preferred action without any intervention. %

\paragraph{Costly interventions.} 
To analyze the effect of the cost of intervention $\cost_\itime$ on the \planner{}'s learnt policy, we assume that the \planner{} decides among three different intervention levels $|\intervention| \in \{0, 0.5, 1\}$ such that $\cost_\itime = |\intervention|$. 
Across different bandit \agent{} tasks $\task^\iagent$ with distinct base rewards $\bm{\reward}^\iagent$ and reward gaps $\suboptgap = \max_{\action\in\actionset}\bm{\reward}^\iagent[\action] - \bm{\reward}^\iagent[\action^*]$, the \planner{} should learn to appropriately incentivize the \agent{} while minimizing the total cost of intervening. 
We then define the experienced reward as: $\forall \action\not= \action^*,(\action,\action^*\in\actionset)$
\eq{
\bm{\realreward}_\itime[\action^*] = \bm{\reward}^\iagent[\action^*] + \intervention_\itime; \quad
\bm{\realreward}_\itime[\action] = \bm{\reward}^\iagent[\action] - \intervention_\itime.
}

Note that this ensures the \agent{} always experiences an intervention, no matter which action it chooses.
During each episode, the \agent{} learns but the \planner{}'s policy is fixed; the \planner{} can update its policy only at the end of each episode (\cref{algo:our_algo}). 
Also, we assume that the \planner{} can only observe the agent's actions $\action^\iagent_\itime$ but not its base reward $\bm{\reward}^\iagent$ or policy update function $\agentlearningalgo^\iagent$.
We measure the performance of the \planner{} using \cref{eq:planner-objective}, with $\df = 1$.

\paragraph{World model.}
The world model predicts the \agent{}'s next \emph{action} (given the \planner{}'s prior observations) to characterize the \agent{}'s behavior. 
We do not train the \planner{}'s world model to estimate the base rewards,
because bandit \agent{}s with distinct base rewards could still execute the same sequence of actions, depending on the agent's explore-exploit algorithm and its observations.

\paragraph{Challenges in the sequential setting.} 
Compared to the setting in \cref{sec:single-shot-setting}, 
learning to intervene on sequential (bandit) learners (\cref{subsec:bandit_descr}) creates more challenges:

    1) Bandit \agent{}s may use different strategies to maximize their experienced reward. 
    The agent's rate of exploration may be constant (e.g., $\epsilon$-greedy) or it can reduce with time (e.g., UCB) \emph{within an episode}. %
    This creates a highly non-stationary environment for the \planner{}: its intervention policy must adapt to different intra-episode explore-exploit behaviors for the same \agent{}. %
    When the \agent{} explores a larger action space, it further exacerbates these challenges %
   as the \planner{} only has partial information about the agent.
    
    2) Bandit \agent{}s are sequential learners and feedback ($\action^\iagent_\itime$, $\realreward^\iagent_\itime$) can update the policy $\policy^\iagent$ differently at different $\itime$.
    This may depend on how optimistic (e.g., UCB) or pessimistic (e.g., EXP3) the \agent{}s are about their reward estimates.
    Hence, an intervention $\action^\iplanner$ may change \agent{} behavior differently at different $\itime$.
    Since interventions have different costs, a strategic \planner{} must decide \emph{when} to intervene and \emph{how much} ($|\intervention|$) based on its observations of the agent's actions.

\cref{subsec:bandit_viz} further illustrates the difficulty of this setting. %

\paragraph{Results.} 

Here, we use 15 bandit \agent{}s for training and 10 bandit \agent{}s for testing, each with different base rewards (both within and across train and test sets). $|\actionset|=10$.
We consider two \agent{} learning algorithms (UCB and $\epsilon$-greedy) %
and a range of exploration vs exploitation characteristics, determined by their exploration coefficients: $\beta \in \{0.17,0.27,0.42,0.5,0.67\}$ for UCB (higher $\beta$ gives more exploration) and $\epsilon\in\{0.1,0.2,0.3,0.4,0.5\}$ for $\epsilon$-greedy (higher $\epsilon$ gives more exploration). 
These constants were chosen such that they yield, on average, the same number of exploratory actions with either UCB or $\epsilon$-greedy without any intervention (see \cref{subsec:designbeta}).%

\cref{tab:results_table} shows the one-shot adapted \planner{}'s score on each test set over $\horizon=200$ time steps.
We compare \oursolution{} with 1) model-free baselines (MF-RL using REINFORCE, MF-MAML using MAML) and 2) REINFORCE with world model (WM-RL); see \cref{subsec:bandit_baselines}. 
We also include zero-shot evaluation of the trained \oursolution{} intervention policy, showing that it outperforms one-shot adapted baselines on unseen test agents. 
We further include a ``No Intervention'' baseline to show how \agent{}s behave by default. 
In all, our results show that \oursolution{}'s model-based meta-learning approach is highly effective: the \planner{} obtains a higher score across \agent{}s with different learning algorithms and explore-exploit behaviors. 

\paragraph{Out-of-distribution performance.} 
\cref{tab:results_table} shows the \planner{}'s score when evaluated on test \agent{}s with the same algorithm but a \emph{different exploration constant} than train \agent{}s. 
Using meta-learning (MF-MAML) and using a world model to predict the \agent{}'s behavior (WM-RL) both have advantages for training a robust and one-shot adaptable intervention policy.
A world model is advantageous when 
1) the test \agent{} is more exploratory than the train set (e.g., $\epsilon=0.1$ at training, $\epsilon=0.4$ at test), or
2) the \agent{} explores throughout an episode and is likely to often select actions other than the one with its current maximum mean reward estimate (e.g., $\epsilon=0.5$ at training).
Because we evaluate on $\testK=1$, fine-tuning on only a single test-time episode, a trained world model provides a useful prior belief for the \planner{}.
Indeed, the MF-RL results show the hidden state representation of the model-free \planner{} might be unable to adapt to high environment non-stationarity without a trained next-\agent{}-action world model. 

Compared to an $\epsilon$-greedy \agent{}, the UCB \agent{} explores mostly at the start of an episode, for all $\beta$.
Hence, with UCB \agent{}s, the \planner{} learns an effective one-shot adaptable intervention policy using meta-learning (MF-MAML) only (even without a world model), as the \agent{}s cause less distribution shift across different $\beta$.
It further emphasizes the effectiveness of meta-learning for %
adaptive policy learning: unlike MF-MAML, neither the world model nor the intervention policy is meta-learned in WM-RL.
Moreover, it also shows that for the same amount of distribution shift (characterized in \cref{subsec:designbeta}), the relative benefit of a world model or meta-learning depends on the nature of the \agent{}'s exploration strategy (which is unknown to the \planner{}). 

\cref{tab:additional_seeds_table}, \cref{app:seeds} includes observations from training with additional random seeds for the experiments with \oursolution{} ($K = 1$) in \cref{tab:results_table}.

\paragraph{Agent exploration vs intervention cost.}
In order to intervene effectively, the \planner{} should learn \emph{when} to intervene and \emph{how much} to incentivize the \agent{} while minimizing its incurred cost.
This is a challenging learning problem for the \planner{} not just during training, but more so during one-shot adaptation at test time.
Bandit algorithms like EXP3 \citep{auer2002nonstochastic} use pessimism in the face of uncertainty, and encourage continued exploration. 
This increases the non-stationarity for the \planner{}.
In order to effectively incentivize such \agent{}s to prefer $\action^*$, the \planner{} needs to accurately predict the \agent{}'s policy from its observations;
otherwise it can incur a high cost for intervening ineffectively \emph{and} lowering its score, and learn to stop intervening. %
Indeed, our results when training on $\epsilon=0.5$-greedy agents show that the MF-RL and MF-MAML \planner{} stop intervening.
In contrast, in that setting, \oursolution{} learns an effective intervention policy that outperforms all baselines, even under distribution shift between train and test agents.

\paragraph{Cross-algorithm evaluation.} 

\cref{tab:zero_shot_cross_table} shows the \planner{}'s scores in the zero-shot generalization setting when the training agent and test agents are of different types (different algorithms and different exploration coefficients). 
We consider only the WM-RL baseline to evaluate the generalization ability of the world model with standard policy gradients vs world model with meta-gradients, without adaptation to unseen test agents. 
We observe that when trained with UCB agents, \oursolution{} outperforms WM-RL for generalizing to $\epsilon$-greedy agents that have a lower exploration coefficient $\epsilon = 0.1$ or $0.2$. 
In contrast, when trained with $\epsilon$-greedy agents, \oursolution{} outperforms WM-RL for generalizing to UCB agents with both higher and lower levels of exploration. 
Note that the behavior of UCB agents is less stochastic than $\epsilon$-greedy agents. 
More generally, a meta-learning \planner{} that is trained on a stochastic agent generalizes well to an equal or less stochastic agent in the zero-shot setting.

\section{\update{Limitations and Future Work}}
\label{sec:discussion}
We have shown that \oursolution{} is an effective framework to learn \planner{} intervention policies that adapt and generalize well to agents with unseen learning behavior. 
\update{
But our focus has been on stateless, sequential adaptive \agent{}s.  
Extending this setup to RL agents that solve non-Markovian settings (e.g., Markov Decision Processes) would introduce a more challenging learning problem for the \planner{} and may require different neural network architectures for the \planner{}'s world model and intervention policy. 
}
Future work could also extend \oursolution{} to settings with multiple learning agents who may coordinate, compete, or a combination thereof.
\update{
While our results show that \oursolution{} can adapt to bandit \agent{}s whose learning algorithm differs between training and test time, we would like to note that the meta-learning framework only guarantees adaptivity within the same meta-distribution over the training and test tasks. 
}
Finally, it would also be interesting to extend \oursolution{} to agents that adapt adversarially to the \planner{}'s intervention policy, which poses a challenging non-stationary problem for the \planner{}.

\bibliography{ref}

\begin{thebibliography}{32}
\providecommand{\natexlab}[1]{#1}
\providecommand{\url}[1]{\texttt{#1}}
\expandafter\ifx\csname urlstyle\endcsname\relax
  \providecommand{\doi}[1]{doi: #1}\else
  \providecommand{\doi}{doi: \begingroup \urlstyle{rm}\Url}\fi

\bibitem[Abbeel \& Ng(2004)Abbeel and Ng]{abbeel2004apprenticeship}
Pieter Abbeel and Andrew~Y Ng.
\newblock Apprenticeship learning via inverse reinforcement learning.
\newblock In \emph{Proceedings of the twenty-first international conference on
  Machine learning}, pp.\ ~1, 2004.

\bibitem[Argall et~al.(2009)Argall, Chernova, Veloso, and
  Browning]{argall2009survey}
Brenna~D Argall, Sonia Chernova, Manuela Veloso, and Brett Browning.
\newblock A survey of robot learning from demonstration.
\newblock \emph{Robotics and autonomous systems}, 57\penalty0 (5):\penalty0
  469--483, 2009.

\bibitem[Auer et~al.(2002)Auer, Cesa-Bianchi, Freund, and
  Schapire]{auer2002nonstochastic}
Peter Auer, Nicolo Cesa-Bianchi, Yoav Freund, and Robert~E Schapire.
\newblock The nonstochastic multiarmed bandit problem.
\newblock \emph{SIAM journal on computing}, 32\penalty0 (1):\penalty0 48--77,
  2002.

\bibitem[Ben-Ayed \& Blair(1990)Ben-Ayed and Blair]{ben1990computational}
Omar Ben-Ayed and Charles~E Blair.
\newblock Computational difficulties of bilevel linear programming.
\newblock \emph{Operations Research}, 38\penalty0 (3):\penalty0 556--560, 1990.

\bibitem[Bengio et~al.(2009)Bengio, Louradour, Collobert, and
  Weston]{bengio2009curriculum}
Yoshua Bengio, J{\'e}r{\^o}me Louradour, Ronan Collobert, and Jason Weston.
\newblock Curriculum learning.
\newblock In \emph{Proceedings of the 26th annual international conference on
  machine learning}, pp.\  41--48, 2009.

\bibitem[Boutilier et~al.(2020)Boutilier, Hsu, Kveton, Mladenov, Szepesvari,
  and Zaheer]{NEURIPS2020_171ae1bb}
Craig Boutilier, Chih-wei Hsu, Branislav Kveton, Martin Mladenov, Csaba
  Szepesvari, and Manzil Zaheer.
\newblock Differentiable meta-learning of bandit policies.
\newblock In H.~Larochelle, M.~Ranzato, R.~Hadsell, M.F. Balcan, and H.~Lin
  (eds.), \emph{Advances in Neural Information Processing Systems}, volume~33,
  pp.\  2122--2134. Curran Associates, Inc., 2020.
\newblock URL
  \url{https://proceedings.neurips.cc/paper/2020/file/171ae1bbb81475eb96287dd78565b38b-Paper.pdf}.

\bibitem[Chen et~al.(2018)Chen, Frazier, and Kempe]{chen2018incentivizing}
Bangrui Chen, Peter Frazier, and David Kempe.
\newblock Incentivizing exploration by heterogeneous users.
\newblock In \emph{Conference On Learning Theory}, pp.\  798--818. PMLR, 2018.

\bibitem[Colson et~al.(2007)Colson, Marcotte, and Savard]{colson2007overview}
Beno{\^\i}t Colson, Patrice Marcotte, and Gilles Savard.
\newblock An overview of bilevel optimization.
\newblock \emph{Annals of operations research}, 153\penalty0 (1):\penalty0
  235--256, 2007.

\bibitem[Eisenhardt(1989)]{eisenhardt1989agency}
Kathleen~M Eisenhardt.
\newblock Agency theory: An assessment and review.
\newblock \emph{Academy of management review}, 14\penalty0 (1):\penalty0
  57--74, 1989.

\bibitem[Finn et~al.(2017{\natexlab{a}})Finn, Abbeel, and
  Levine]{finn2017model}
Chelsea Finn, Pieter Abbeel, and Sergey Levine.
\newblock Model-agnostic meta-learning for fast adaptation of deep networks.
\newblock In \emph{International conference on machine learning}, pp.\
  1126--1135. PMLR, 2017{\natexlab{a}}.

\bibitem[Finn et~al.(2017{\natexlab{b}})Finn, Abbeel, and
  Levine]{finn_model-agnostic_2017}
Chelsea Finn, Pieter Abbeel, and Sergey Levine.
\newblock Model-{Agnostic} {Meta}-{Learning} for {Fast} {Adaptation} of {Deep}
  {Networks}.
\newblock \emph{arXiv:1703.03400 [cs]}, July 2017{\natexlab{b}}.
\newblock URL \url{http://arxiv.org/abs/1703.03400}.

\bibitem[Foerster et~al.(2018)Foerster, Chen, Al-Shedivat, Whiteson, Abbeel,
  and Mordatch]{foerster_learning_2018}
Jakob~N. Foerster, Richard~Y. Chen, Maruan Al-Shedivat, Shimon Whiteson, Pieter
  Abbeel, and Igor Mordatch.
\newblock Learning with {Opponent}-{Learning} {Awareness}.
\newblock \emph{arXiv:1709.04326 [cs]}, September 2018.
\newblock URL \url{http://arxiv.org/abs/1709.04326}.

\bibitem[Guo et~al.(2021)Guo, Agrawal, Grover, Muthukumar, and
  Pananjady]{https://doi.org/10.48550/arxiv.2106.14866}
Wenshuo Guo, Kumar~Krishna Agrawal, Aditya Grover, Vidya Muthukumar, and Ashwin
  Pananjady.
\newblock Learning from an exploring demonstrator: Optimal reward estimation
  for bandits, 2021.
\newblock URL \url{https://arxiv.org/abs/2106.14866}.

\bibitem[Hurwicz \& Reiter(2006)Hurwicz and Reiter]{hurwicz2006designing}
Leonid Hurwicz and Stanley Reiter.
\newblock \emph{Designing economic mechanisms}.
\newblock Cambridge University Press, 2006.

\bibitem[Jacq et~al.(2019)Jacq, Geist, Paiva, and Pietquin]{jacq2019learning}
Alexis Jacq, Matthieu Geist, Ana Paiva, and Olivier Pietquin.
\newblock Learning from a learner.
\newblock In \emph{International Conference on Machine Learning}, pp.\
  2990--2999. PMLR, 2019.

\bibitem[Kingma \& Ba(2014)Kingma and Ba]{kingma2014adam}
Diederik~P Kingma and Jimmy Ba.
\newblock Adam: A method for stochastic optimization.
\newblock \emph{arXiv preprint arXiv:1412.6980}, 2014.

\bibitem[Lai et~al.(1985)Lai, Robbins, et~al.]{lai1985asymptotically}
Tze~Leung Lai, Herbert Robbins, et~al.
\newblock Asymptotically efficient adaptive allocation rules.
\newblock \emph{Advances in applied mathematics}, 6\penalty0 (1):\penalty0
  4--22, 1985.

\bibitem[Lowe et~al.(2017)Lowe, WU, Tamar, Harb, Pieter~Abbeel, and
  Mordatch]{lowe_multi-agent_2017}
Ryan Lowe, YI~WU, Aviv Tamar, Jean Harb, OpenAI Pieter~Abbeel, and Igor
  Mordatch.
\newblock Multi-{Agent} {Actor}-{Critic} for {Mixed}
  {Cooperative}-{Competitive} {Environments}.
\newblock In I.~Guyon, U.~V. Luxburg, S.~Bengio, H.~Wallach, R.~Fergus,
  S.~Vishwanathan, and R.~Garnett (eds.), \emph{Advances in {Neural}
  {Information} {Processing} {Systems} 30}, pp.\  6379--6390. Curran
  Associates, Inc., 2017.

\bibitem[Luketina et~al.(2022)Luketina, Flennerhag, Schroecker, Abel, Zahavy,
  and Singh]{luketina2022meta}
Jelena Luketina, Sebastian Flennerhag, Yannick Schroecker, David Abel, Tom
  Zahavy, and Satinder Singh.
\newblock Meta-gradients in non-stationary environments.
\newblock In \emph{ICLR Workshop on Agent Learning in Open-Endedness}, 2022.

\bibitem[Maghsudi et~al.(2021)Maghsudi, Lan, Xu, and van
  Der~Schaar]{maghsudi2021personalized}
Setareh Maghsudi, Andrew Lan, Jie Xu, and Mihaela van Der~Schaar.
\newblock Personalized education in the artificial intelligence era: what to
  expect next.
\newblock \emph{IEEE Signal Processing Magazine}, 38\penalty0 (3):\penalty0
  37--50, 2021.

\bibitem[Milgrom \& Milgrom(2004)Milgrom and Milgrom]{milgrom2004putting}
Paul Milgrom and Paul~Robert Milgrom.
\newblock \emph{Putting auction theory to work}.
\newblock Cambridge University Press, 2004.

\bibitem[Nagabandi et~al.(2018)Nagabandi, Clavera, Liu, Fearing, Abbeel,
  Levine, and Finn]{nagabandi2018learning}
Anusha Nagabandi, Ignasi Clavera, Simin Liu, Ronald~S Fearing, Pieter Abbeel,
  Sergey Levine, and Chelsea Finn.
\newblock Learning to adapt in dynamic, real-world environments through
  meta-reinforcement learning.
\newblock In \emph{International Conference on Learning Representations}, 2018.

\bibitem[Pardoe et~al.(2006)Pardoe, Stone, Saar-Tsechansky, and
  Tomak]{10.1145/1151454.1151480}
David Pardoe, Peter Stone, Maytal Saar-Tsechansky, and Kerem Tomak.
\newblock Adaptive mechanism design: A metalearning approach.
\newblock In \emph{Proceedings of the 8th International Conference on
  Electronic Commerce: The New e-Commerce: Innovations for Conquering Current
  Barriers, Obstacles and Limitations to Conducting Successful Business on the
  Internet}, ICEC '06, pp.\  92–102, New York, NY, USA, 2006. Association for
  Computing Machinery.
\newblock ISBN 1595933921.
\newblock \doi{10.1145/1151454.1151480}.
\newblock URL \url{https://doi.org/10.1145/1151454.1151480}.

\bibitem[Ramponi et~al.(2020)Ramponi, Drappo, and Restelli]{ramponi2020inverse}
Giorgia Ramponi, Gianluca Drappo, and Marcello Restelli.
\newblock Inverse reinforcement learning from a gradient-based learner.
\newblock \emph{Advances in Neural Information Processing Systems},
  33:\penalty0 2458--2468, 2020.

\bibitem[Shi et~al.(2021)Shi, Xu, Xiong, and Shen]{shi2021almost}
Chengshuai Shi, Haifeng Xu, Wei Xiong, and Cong Shen.
\newblock (almost) free incentivized exploration from decentralized learning
  agents.
\newblock \emph{Advances in Neural Information Processing Systems},
  34:\penalty0 560--571, 2021.

\bibitem[Sinha et~al.(2017)Sinha, Malo, and Deb]{sinha2017review}
Ankur Sinha, Pekka Malo, and Kalyanmoy Deb.
\newblock A review on bilevel optimization: from classical to evolutionary
  approaches and applications.
\newblock \emph{IEEE Transactions on Evolutionary Computation}, 22\penalty0
  (2):\penalty0 276--295, 2017.

\bibitem[Sutton \& Barto(1998)Sutton and Barto]{sutton1998introduction}
Richard~S Sutton and Andrew~G Barto.
\newblock Introduction to reinforcement learning, 1998.

\bibitem[Von~Stackelberg(2010)]{von2010market}
Heinrich Von~Stackelberg.
\newblock \emph{Market structure and equilibrium}.
\newblock Springer Science \& Business Media, 2010.

\bibitem[Wang et~al.(2021)Wang, Xu, Li, Liu, and Wang]{wang2021incentivizing}
Huazheng Wang, Haifeng Xu, Chuanhao Li, Zhiyuan Liu, and Hongning Wang.
\newblock Incentivizing exploration in linear bandits under information gap.
\newblock \emph{arXiv preprint arXiv:2104.03860}, 2021.

\bibitem[Williams(1992)]{williams1992simple}
Ronald~J Williams.
\newblock Simple statistical gradient-following algorithms for connectionist
  reinforcement learning.
\newblock \emph{Machine learning}, 8\penalty0 (3):\penalty0 229--256, 1992.

\bibitem[Zheng et~al.(2020)Zheng, Trott, Srinivasa, Naik, Gruesbeck, Parkes,
  and Socher]{zheng_ai_2020}
Stephan Zheng, Alexander Trott, Sunil Srinivasa, Nikhil Naik, Melvin Gruesbeck,
  David~C. Parkes, and Richard Socher.
\newblock The {AI} {Economist}: {Improving} {Equality} and {Productivity} with
  {AI}-{Driven} {Tax} {Policies}.
\newblock \emph{arXiv:2004.13332 [cs, econ, q-fin, stat]}, April 2020.
\newblock URL \url{http://arxiv.org/abs/2004.13332}.

\bibitem[Zheng et~al.(2022)Zheng, Trott, Srinivasa, Parkes, and
  Socher]{doi:10.1126/sciadv.abk2607}
Stephan Zheng, Alexander Trott, Sunil Srinivasa, David~C. Parkes, and Richard
  Socher.
\newblock The ai economist: Taxation policy design via two-level deep
  multiagent reinforcement learning.
\newblock \emph{Science Advances}, 8\penalty0 (18):\penalty0 eabk2607, 2022.
\newblock \doi{10.1126/sciadv.abk2607}.
\newblock URL \url{https://www.science.org/doi/abs/10.1126/sciadv.abk2607}.

\end{thebibliography}
\bibliographystyle{tmlr}

\clearpage

\appendix
\section{Notation}
\label{sec:appendix_notation}
\begin{table}[t]
    \centering
    \caption{\textbf{Overview of notation.}}%
    
    \begin{tabular}{p{0.49\linewidth}p{0.5\linewidth}}
        \textbf{Variable} & \textbf{Symbol} \\
        Time & $\itime$ \\
        \capsPlanner{} & $\iplanner{}$ \\
        Agent & $\iagent{}$\\
        State & $\state$ \\
        State vector & $\statevec$\\
        State space & $\stateset{}$ \\
        Agent's action space & $\actionset$\\
        \capsPlanner{}'s action space & $\actionset^\iplanner$\\
        Action sequence & $\action_{1:\horizon} = \brckcur{\action_1, \action_2,\dots, \action_\horizon}$\\
        Agent $\iagent$'s reward sequence & $\realreward^\iagent_{1:\horizon} = \brckcur{\realreward^\iagent_1,\dots, \realreward^\iagent_\horizon}$ \\
        \capsPlanner{}'s reward sequence & $\reward^\iplanner_{1:\horizon} = \brckcur{\reward^\iplanner_1, \dots, \reward^\iplanner_\horizon}$ \\
        Transition function & $\transition$ \\
        Agent $\iagent$'s policy & $\policy^\iagent$\\
        \capsPlanner{}'s intervention policy  & $\policy^\iplanner$\\
        Agent's mean estimate of intervened rewards for action $\action$ & $\meanadj_\action$\\
        Number of adaptation steps & $\numoptsteps$ \\
        Number of meta-tasks for the planner & $\numtasks$\\
        \capsPlanner{}'s history of interventions and observed agent actions upto time $\itime$ & $\hist^\iplanner_\itime = \brckcur{\action^\iplanner_1,\action^\iagent_1,\action^\iplanner_2,\action^\iagent_2,\dots,\action^\iplanner_{\itime-1},\action^\iagent_{\itime-1}}$ \\
        Agent's history of actions taken and rewards observed upto time $\itime$ & $\hist^\iagent_\itime = \brckcur{\action^\iagent_1,\realreward^\iagent_1,\action^\iagent_2,\realreward^\iagent_2,\dots,\action^\iagent_{\itime-1},\realreward^\iagent_{\itime-1}}$\\
    \end{tabular}
    \label{tab:notation}
\end{table}

\begin{table}[t]
    \caption{\textbf{Notation for \oursolution{}} See Section \cref{sec:our_algo} for their use.}
    \centering
    \begin{tabular}{p{0.49\linewidth}p{0.5\linewidth}}
    \capsPlanner{}'s policy parameter & $\ppolicypar \in \Theta$\\
    Agent $\iagent$'s learning algorithm & $\agentalgo^i \in \mathcal{F}$\\
    Agent $\iagent$'s true action mean rewards & $\mean^\iagent\sim\mathcal{U}$\\
    Agent $\iagent$'s intervened action mean rewards & $\meanadj^\iagent$\\
    \capsPlanner{}'s action at time $\itime$ & $\action^\iplanner_t \sim \policy^\iplanner_\ppolicypar\brck{\action^\iplanner_\itime|\action^\iagent_{\itime-1},\action^\iplanner_{\itime-1},\hat{\action}^\iagent_\itime,\hidden^\iplanner_{\itime-1}}$ \\
    Hidden state space of the \planner{}'s recurrent world model  & $\hiddenstateclass$\\
    Agent's action at time $\itime$  & $\action^\iagent_\itime\sim \policy^\iagent_\itime\brck{\action^\iagent_\itime|\hist^\iagent_\itime}, \itime=1,\dots,\horizon$ \\
    Agent's reward at time $\itime$ & $\reward^\iagent_\itime\sim\mathcal{N}\brck{\meanadj^\iagent,\sigma^2}$ \\
    \capsPlanner{}'s world model estimate of the agent's action probability distribution & $\hat{\policy}^\iagent_\apolicypar:\actionset\times\actionset^\iplanner\times\hiddenstateclass\rightarrow\Delta\brck{\actionset},\;\hat{\policy}^\iagent_{\apolicypar,0}\in\actionset$ \\
    \capsPlanner{}'s world model estimate of the latent state of the environment & $\latentfunc^\iagent_\apolicypar:\actionset\times\actionset^\iplanner\times\hiddenstateclass\rightarrow\hiddenstateclass,\;\latentfunc^\iagent_{\apolicypar,0}\in\Delta\brck{\hiddenstateclass}$ \\
    \capsPlanner{}'s world model hidden state embeding in the LSTM architecture & $\latentpar^\iagent_{\itime} = \latentfunc^\iagent_\apolicypar\brck{\action^\iagent_{\itime-1},\action^\iplanner_{\itime-1},\latentpar_{\itime-1}},\;\itime=2,\dots,\horizon\;\latentpar^\iagent_1=\latentfunc^\iagent_{\apolicypar,0}$
    \end{tabular}
    \label{tab:proof_notation}
\end{table}

For an overview of all symbols and variables used in this work, see \cref{tab:notation} and \cref{tab:proof_notation}.
\section{Additional Results}
\label{app:results}

\subsection{Description of the bandit algorithms}
\label{subsec:bandit_descr}

We provide a brief overview of the learning algorithms referred to in \cref{sec:bandit-setting}.

\paragraph{UCB.} 
This is an Upper Confidence Bound based exploration-exploitation algorithm that follows the principle of optimism in the face of uncertainty. 
At each time step $\itime$, the bandit agent selects an action
\eq{
\action_\itime = \argmax_{\action} \meanadj_\action + \beta\sqrt{\frac{\log \itime}{n_\action}}
}
where $n_\action$ is the number of steps until $\itime$ in which it previously selected the action $\action$, $\meanadj_\action$ is its corresponding mean estimate for the experienced rewards $\realreward$ for action $\action$ and $\beta$ is the exploration constant that balances the amount of exploration vs. exploitation across a time horizon $\horizon$. 
A higher value of $\beta$ makes the agent less optimistic and explore its action space more. 
The UCB agent's tendency to explore is also affected by the difference in the mean reward estimates
of its actions. 
In the context of our \planner{} - agent problem formulation, if the UCB agent has a larger value of $\delta = \max_\action \reward_\action - \reward_{\action^*}$, without any intervention at the beginning of an episode, its confidence bounds would quickly converge to exploiting the action $\argmax_\action \reward_\action$.
So a \planner{} that intervenes only towards the later stages of an episode with this agent would have to provide much more incentives (higher $\intervention$) to alter the agent's preferred action to be $\action^*$, thus incurring a larger cost $\cost$ as compared to a \planner{} that intervenes more at the beginning of an episode when the UCB agent is still exploring its action space. 
This is also illustrated in \cref{sec:single-shot-setting} with a simpler best response agent in the single round game setting. 
As shown in \cref{fig:2x2_u_0.21} and \cref{fig:2x2_u_0.07}, under observation noise (partial information), the meta-trained \planner{} has a better one-shot meta-test-time performance when the agent's base payoff has a higher difference between the \planner{}'s preferred action and the agent's intrinsic preference without any intervention.
Additionally, \cref{subsec:bandit_viz} provides an illustration of this behavior.

\paragraph{$\epsilon$-greedy.}
A simple exploration-exploitation strategy in the bandit setting is the $\epsilon$-greedy rule \citep{sutton1998introduction} wherein the agent selects with probability $1 - \epsilon$ the action $\action_\itime = \argmax_\action \meanadj_\action$ and with probability $\epsilon$ it selects a random action. 
In our setting, we consider $\epsilon$ to be constant during an episode, which results in a uniform exploration rate throughout. 
In contrast to the UCB agent, the $\epsilon$-greedy algorithm simulates a less optimistic, more exploratory agent for which the \planner{} requires a robust belief representation of the agent's predicted behavior conditioned on the \planner{}'s past observations (\cref{tab:results_table}).
Since there is a uniform exploration rate for the agent, the \planner{} has to continue intervening intermittently throughout an episode, especially when $\delta$ is large and the agent could obtain a higher reward for an action $\action \not= \action^*$ by exploring its action space when the \planner{} does not intervene.

\paragraph{EXP3.}
The Exponential-weight algorithm for Exploration and Exploitation (EXP3) \citep{auer2002nonstochastic} follows a more pessimistic approach to exploration-exploitation in the bandit setting.
It maintains a set of weights for each agent action $\action \in \actionset$ which are updated using the experienced rewards $\realreward$ as follows:
\eq{
\policy_\itime(\action_\itime) = \frac{w}{|{\actionset}|} + (1 - w)\frac{\eta \exp\brck{S_{\action_\itime,\itime}}}{\sum_{\action_\itime \in |\actionset|} \eta \exp\brck{S_{\action_\itime,\itime}}}, 
} 
where 
\eq{
S_{\action_\itime,\itime} = \sum_{l=1}^\itime \mathbf{1}\brckcur{\action_l = \action_\itime}\frac{\realreward_{\action_\itime,l}}{\policy_l},\;\eta=\frac{w}{|\actionset|}.
}
Here, $w$ is the variable that determines the extent of uniform random exploration in the action space.
This presents a very challenging problem to learn a suitable belief representation for such agents that can be utilized by a \planner{} to guide its intervention policy. 
In \cref{sec:bandit-setting}, we exclude EXP3 from \cref{tab:results_table} since it is primarily designed for an adversarial bandit setup, whereas we do not consider an agent to have such biases under our current problem formulation.

\subsection{Characterizing the distribution shift in our evaluation setup}
\label{subsec:designbeta}

\begin{table}[t]
\caption{\textbf{Experiment design choice.} Frequency of agent selecting $\action_\itime \not= \argmax_\action \reward_\action$ with UCB and $\epsilon$-greedy algorithms on the same set of base rewards (without any intervention) with a horizon $\horizon=200$, averaged across 3 random seeds.}
    \centering
    \begin{tabular}{lc|cr}%
    $\beta$ & UCB & $\epsilon$-greedy & $\epsilon$\\
    \toprule
    0.17 & 33 (0) & 33 (0) & 0.10\\
    0.27 & 47 (0) & 47 (4) & 0.20\\
    0.42 & 70 (0) & 68 (9) & 0.30\\
    0.50 & 80 (0) & 81 (3) & 0.40\\
    0.67 & 99 (0) & 99 (1) & 0.50
    \end{tabular}
    \label{tab:Tab_distr_shift}
\end{table}

Bandit agents having the same base reward $\reward$ make different explore-exploit decisions depending on their algorithm (eg. UCB, $\epsilon$-greedy) and also their prior observations.
In \cref{sec:bandit-setting}, we consider agents with the same set of base rewards, but following different bandit algorithms.
Both UCB and $\epsilon$-greedy have tunable parameters that determine their explore-exploit tradeoff.
In order to measure the robustness of the learnt \planner{} policy to different agent behavior (leading to different levels of non-stationarity in the \planner{}'s environment between training and test agents), we vary the amount of exploration performed by the agent by varying the respective parameters: $\beta$ for the UCB agent and $\epsilon$ for the $\epsilon$-greedy agent.
\cref{tab:Tab_distr_shift} shows the average (and standard error) frequency of exploration by the agents for our choices of $\beta$ and $\epsilon$ in \cref{sec:bandit-setting}.
We vary $\beta$ and $\epsilon$ such that they are pairwise comparable in \cref{tab:results_table} and would lead to similar change in exploration frequency for both UCB and $\epsilon$-greedy agents.
In other words, following \cref{tab:results_table}, a \planner{} trained with UCB agents having $\beta=0.17$ when evaluated with UCB agents having $\beta\in\{0.17,0.27,0.42,0.50,0.67\}$  will encounter a similar shift in the agent's exploration frequency as in the case of training with $\epsilon$-greedy agents with $\epsilon=0.1$ and evaluating on $\epsilon$-greedy agents having $\epsilon\in\{0.1,0.2,0.3,0.4,0.5\}$.
In that case, the difference in achieved scores between the UCB and $\epsilon$-greedy agents can be attributed to the way in which they distribute their exploratory actions: UCB agent being more optimistic focuses most of its exploration at the beginning of an episode, whereas the $\epsilon$-greedy agent is more stochastic with uniform random exploration throughout.

\subsection{Visualizing the effect of $\beta$ and the effect of different types of \planner{}'s interventions on the behavior of a UCB agent}
\label{subsec:bandit_viz}

We will consider three different instances of base rewards for a UCB agent and characterize its behavior when 
\begin{itemize}
    \item unintervened
    \item the \planner{} intervenes once every 10 time steps for $\horizon=200$ (strategy S1)
    \item the \planner{} intervenes continuously until the first 20 time steps for $\horizon=200$ (strategy S2)
\end{itemize} for $\beta\in\{0.17,0.27,0.42,0.50,0.67\}$. 
Instead of a learned stochastic intervention policy, we will analyze the effect of the deterministic policies S1 and S2 in aligning the preferred action of an \agent{} with the preferred action of the \planner{}.
Note that both S1 and S2 incur the same total intervention cost. 

\begin{figure}[htp]
    \centering
    \begin{subfigure}{0.5\linewidth}
        \centering
       \includegraphics[scale=0.35]{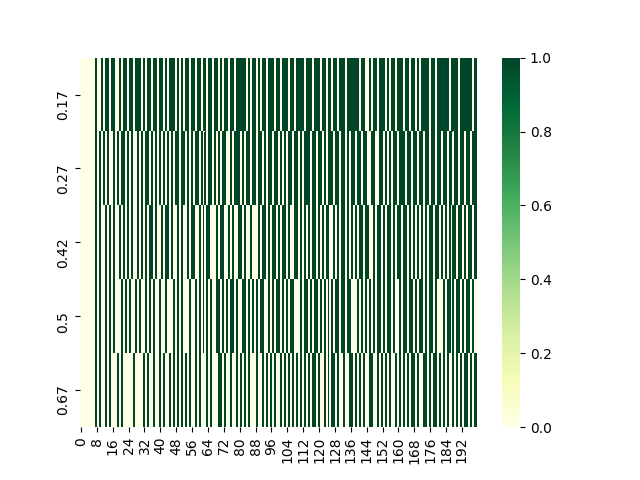}
        \caption{}
        \label{fig:viz_ag6_convergence}
    \end{subfigure}%
    \begin{subfigure}{0.5\linewidth}
        \centering
       \includegraphics[scale=0.35]{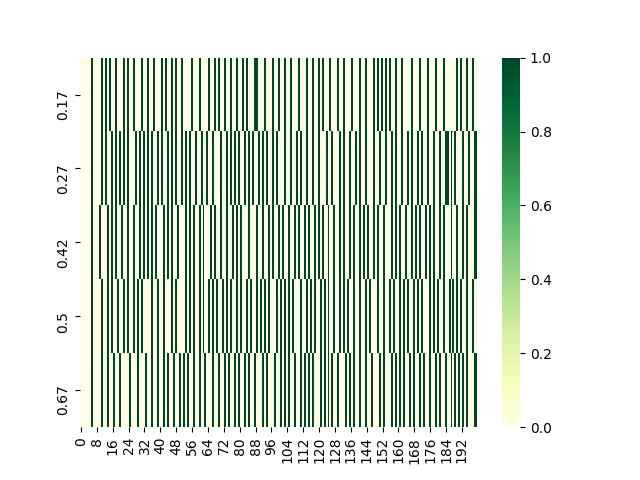}
        \caption{}
        \label{fig:viz_ag6_no_interv}
    \end{subfigure}
    \begin{subfigure}{0.5\linewidth}
    \centering
       \includegraphics[scale=0.35]{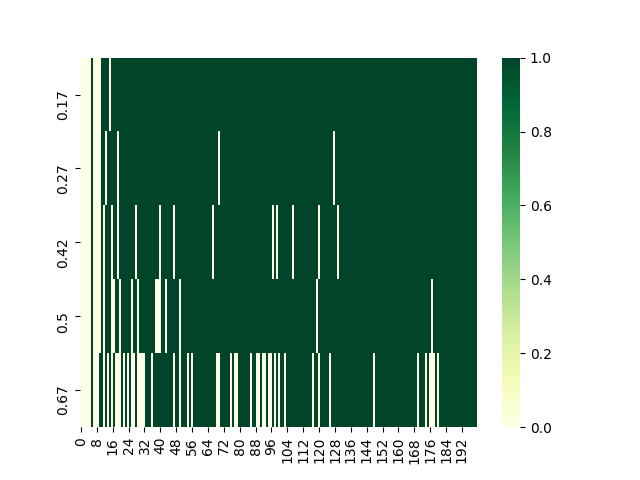}
        \caption{}
        \label{fig:viz_ag6_interv10}
    \end{subfigure}%
    \begin{subfigure}{0.5\linewidth}
    \centering
       \includegraphics[scale=0.35]{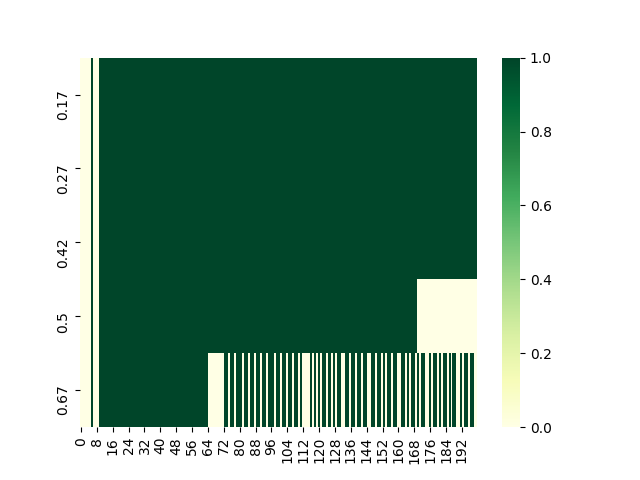}
        \caption{}
        \label{fig:viz_ag6_intervT20}
    \end{subfigure}
    \caption{\textbf{Characterizing \agent{}'s behavior.} UCB agent with base rewards $\brcksq{0.16, 0.11, 0.66, 0.14, 0.20, 0.37, \textbf{0.82}, 0.10, \textbf{0.84}, 0.10}$. The agent prefers the action with base reward 0.84, while the \planner{} prefers the action with base reward 0.82. Horizontal axis indicates time steps $t = \{1,\dots,200\}$. Vertical axis indicates agents following UCB with different exploration coefficient $\beta$. Values are either 0 or 1. (a) Frequency distribution of \agent{} selecting its unintervened preferred action with base reward 0.84. (b) Frequency distribution of \agent{} selecting $\action^*$ without \planner{}'s intervention. (c) Frequency distribution of \agent{} selecting $\action^*$ under \planner{}'s intervention S1. (d) Frequency distribution of \agent{} selecting $\action^*$ under \planner{}'s intervention S2. For a small $\suboptgap = \max_{\action\in\actionset}\bm{\reward}^\iagent[\action] - \bm{\reward}^\iagent[\action^*] = 0.02$, both S1 and S2 affect the agent's behavior quite similarly.}
    \label{fig:UCB_viz_ag6}
\end{figure}

First, consider a UCB agent whose base reward is $\bm\reward = \brcksq{0.16, 0.11, 0.66, 0.14, 0.20, 0.37, \textbf{0.82}, 0.10, \textbf{0.84}, 0.10}$ (\cref{fig:UCB_viz_ag6}). 
The \planner{} prefers the action with base reward 0.82, whereas the unintervened \agent{} would prefer the action with base reward 0.84.
For such small value of $\suboptgap = \max_{\action\in\actionset}\bm{\reward}^\iagent[\action] - \bm{\reward}^\iagent[\action^*] = 0.02$, the agent can be incentivized to align its preferred action with $\action^*$ more easily than if $\suboptgap$ were larger.  
\cref{fig:UCB_viz_ag6} depicts the behavior of the UCB agents with different exploration coefficients $\beta$ under different \planner{}-\agent{} interaction conditions over $\horizon=200$ as follows.
In \cref{fig:viz_ag6_convergence}, a value of 1 indicates the time step when the unintervened agent selects the action with base reward 0.84, whereas \cref{fig:viz_ag6_no_interv} similarly shows its frequency of selecting $\action^*$ without the \planner{}'s intervention.
As expected, these frequency distributions are quite similar since $\suboptgap = 0.02$.
Given the same budget for interventions, \cref{fig:viz_ag6_interv10} shows the agent's response to the \planner{}'s intervention strategy S1 and \cref{fig:viz_ag6_intervT20} shows its response to the intervention strategy S2. 
In these figures, a value of 1 indicates the time step when the agent selects $\action^*$. 
Both the \planner{} intervention policies are able to align the action preference of the UCB agent with the \planner{}'s preference. 
But S2 receives a higher score, especially for agents with lower values for $\beta$. 
This is because the UCB algorithm gradually shifts from exploration to exploitation as the episode progresses, and since S2 uses its intervention budget in the initial 20 time steps of the episode, the \planner{} is able to incentivize the agent by effectively changing the observed reward $\realreward$ at the beginning of the episode. 
It experimentally verifies our discussion in \cref{sec:bandit-setting}: in sequential learners, the time step \emph{when} the \planner{} intervenes determines how effective the intervention will be in helping align the \agent{}'s action preference with that of the \planner{}. 
This becomes more pronounced for higher values of $\suboptgap$, which we analyze next.

\begin{figure}[htp]
    \centering
    \begin{subfigure}{0.5\linewidth}
    \centering
       \includegraphics[scale=0.35]{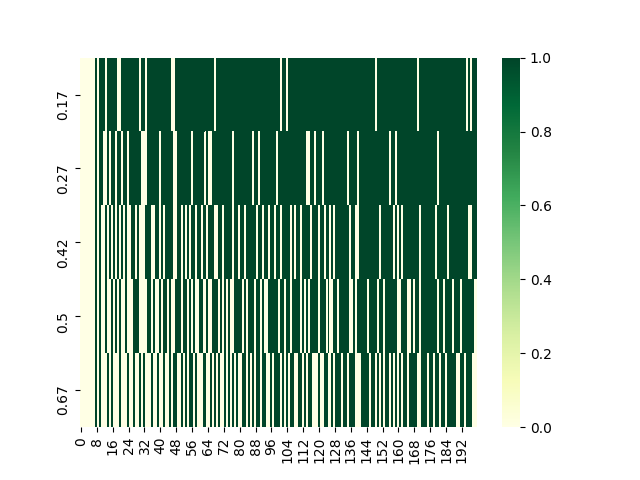}
        \caption{}
        \label{fig:viz_ag9_convergence}
    \end{subfigure}%
    \begin{subfigure}{0.5\linewidth}
    \centering
       \includegraphics[scale=0.35]{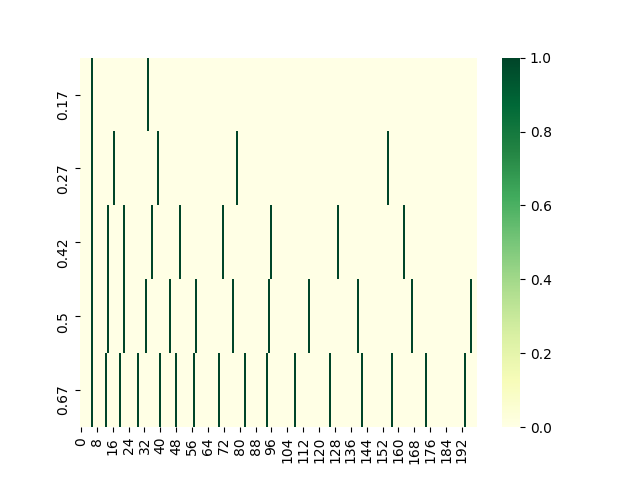}
        \caption{}
        \label{fig:viz_ag9_no_interv}
    \end{subfigure}
    \begin{subfigure}{0.5\linewidth}
    \centering
       \includegraphics[scale=0.35]{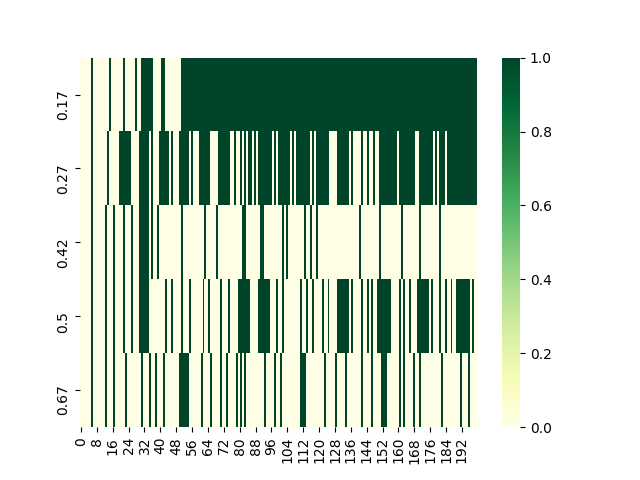}
        \caption{}
        \label{fig:viz_ag9_interv10}
    \end{subfigure}%
    \begin{subfigure}{0.5\linewidth}
    \centering
       \includegraphics[scale=0.35]{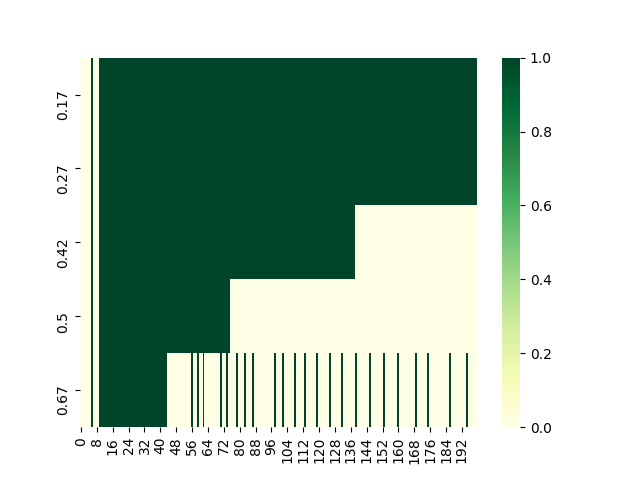}
        \caption{}
        \label{fig:viz_ag9_intervT20}
    \end{subfigure}
    \caption{\textbf{Characterizing \agent{}'s behavior.} UCB agent with base rewards $\brcksq{0.32, 0.67, 0.13, 0.72, 0.29, 0.18, \textbf{0.59}, 0.02, \textbf{0.83}, 0.01}$. The agent prefers the action with base reward 0.83, while the \planner{} prefers the action with base reward 0.59. Horizontal axis indicates time steps $t = \{1,\dots,200\}$. Vertical axis indicates agents following UCB with different exploration coefficient $\beta$. Values are either 0 or 1. (a) Frequency distribution of \agent{} selecting its unintervened preferred action with base reward 0.83. (b) Frequency distribution of \agent{} selecting $\action^*$ without \planner{}'s intervention. (c) Frequency distribution of \agent{} selecting $\action^*$ under \planner{}'s intervention S1. (d) Frequency distribution of \agent{} selecting $\action^*$ under \planner{}'s intervention S2. For different values of $\beta$, the UCB agent acts differently based on \emph{when} the \planner{} intervened following S1 or S2. S1 intervenes periodically whereas S2 intervenes only at the beginning. The action selected by the agents in an episode clearly reflects the effect that this has in being able to align the agent's preference with that of the \planner{}.}
    \label{fig:UCB_viz_ag9}
\end{figure}

\cref{fig:UCB_viz_ag9} demonstrates the behavior of a UCB agent with base reward $\bm\reward = \brcksq{0.32, 0.67, 0.13, 0.72, 0.29, 0.18, \textbf{0.59}, 0.02, \textbf{0.83}, 0.01}$. 
The \planner{} prefers the action with base reward 0.59. 
Without any interventions, the agent prefers the action with base reward 0.83.
In this case, $\suboptgap = \max_{\action\in\actionset}\bm{\reward}^\iagent[\action] - \bm{\reward}^\iagent[\action^*] = 0.24$. 
\cref{fig:viz_ag9_no_interv} indicates the frequency with which the unintervened agent selects $\action^*$ whereas \cref{fig:viz_ag9_convergence} shows the frequency of selecting the action with base reward 0.83 without the \planner{}'s intervention. 
Note that the unintervened agent would rarely pick $\action^*$, even less so for smaller values of the exploration coefficient $\beta$.
\cref{fig:viz_ag9_interv10} indicates the frequency of the agent selecting $\action^*$ with \planner{}'s intervention policy S1 and \cref{fig:viz_ag9_intervT20} indicates its frequency of selection $\action^*$ with \planner{}'s intervention policy S2. 
We observe that S2 outperforms S1 in aligning the agent's preferred action with the \planner{}'s preferred action. 
Since UCB agents tend to explore their action space more at the beginning of the episode, intervening on the agent's experienced reward during the initial time steps (S2) has a more noticeable effect in influencing the agent's preferred action than intervening with a fixed interval (S1).

\begin{figure}[htp]
    \centering
    \begin{subfigure}{0.5\linewidth}
    \centering
       \includegraphics[scale=0.35]{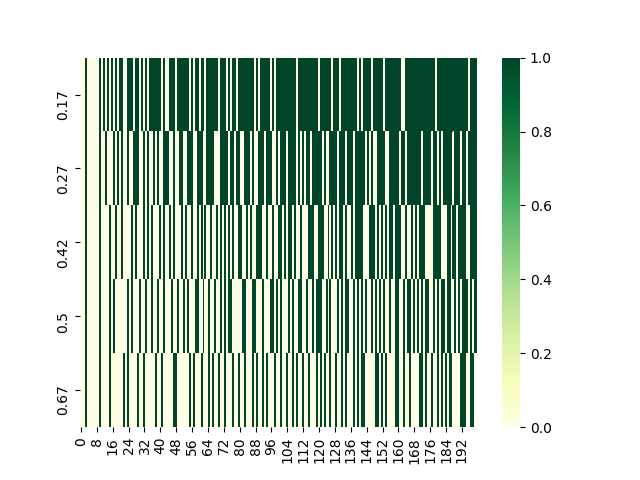}
        \caption{}
        \label{fig:viz_ag1_convergence}
    \end{subfigure}%
    \begin{subfigure}{0.5\linewidth}
    \centering
       \includegraphics[scale=0.35]{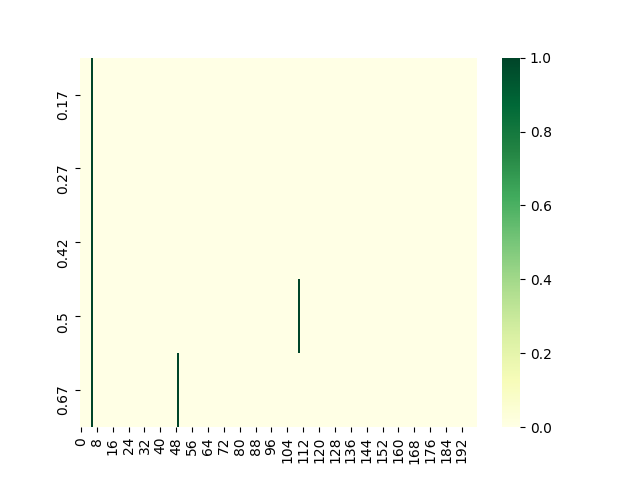}
        \caption{}
        \label{fig:viz_ag1_no_interv}
    \end{subfigure}
    \begin{subfigure}{0.5\linewidth}
    \centering
       \includegraphics[scale=0.35]{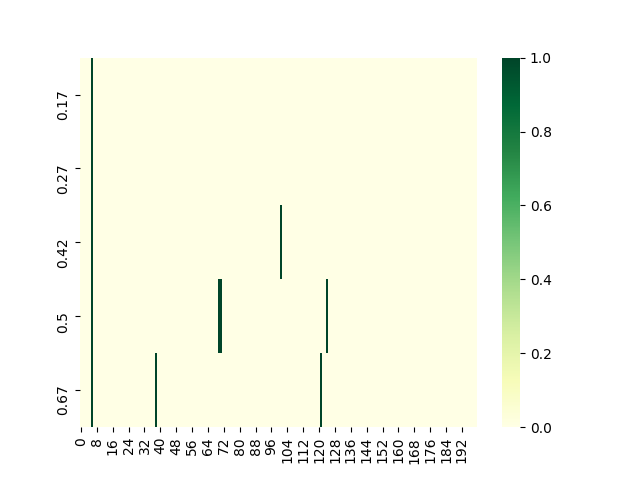}
        \caption{}
        \label{fig:viz_ag1_interv10}
    \end{subfigure}%
    \begin{subfigure}{0.5\linewidth}
    \centering
       \includegraphics[scale=0.35]{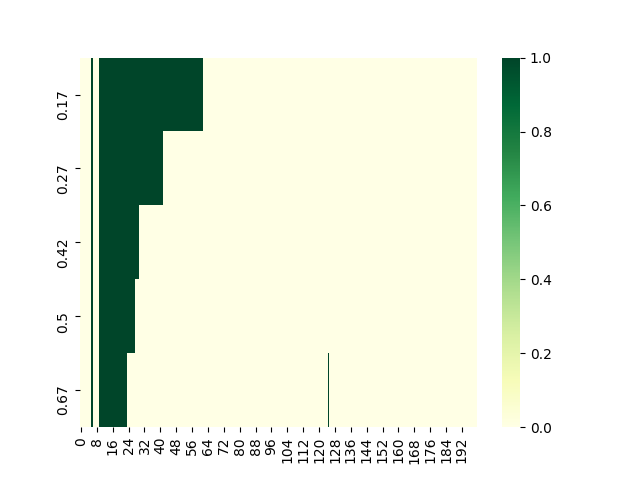}
        \caption{}
        \label{fig:viz_ag1_intervT20}
    \end{subfigure}
    \caption{\textbf{Characterizing \agent{}'s behavior.} UCB agent with base rewards $\brcksq{0.79, 0.53, 0.57, \textbf{0.93}, 0.07, 0.09, \textbf{0.02}, 0.83, 0.78, 0.87}$. The agent prefers the action with base reward 0.93, while the \planner{} prefers the action with base reward 0.02. Horizontal axis indicates time steps $t = \{1,\dots,200\}$. Vertical axis indicates agents following UCB with different exploration coefficient $\beta$. Values are either 0 or 1. (a) Frequency distribution of \agent{} selecting its unintervened preferred action with base reward 0.93. (b) Frequency distribution of \agent{} selecting $\action^*$ without \planner{}'s intervention. (c) Frequency distribution of \agent{} selecting $\action^*$ under \planner{}'s intervention S1. (d) Frequency distribution of \agent{} selecting $\action^*$ under \planner{}'s intervention S2. Since $\suboptgap = \max_{\action\in\actionset}\bm{\reward}^\iagent[\action] - \bm{\reward}^\iagent[\action^*] = 0.91$ is large and the UCB agent explores its action space in the initial time steps, S2 is able to intervene more effectively than S1 and achieves a higher score. Since the agent is a sequential learner, it discovers its own preferred action once the \planner{} stops intervening and does not select $\action^*$.}
    \label{fig:UCB_viz_ag1}
\end{figure}

Similar observations also hold in \cref{fig:UCB_viz_ag1} where the UCB agent has a base reward $\bm\reward = \brcksq{0.79, 0.53, 0.57, \textbf{0.93}, 0.07, 0.09, \textbf{0.02}, 0.83, 0.78, 0.87}$ and the \planner{} prefers the action with base reward $0.02$. 
In this case, $\suboptgap = \max_{\action\in\actionset}\bm{\reward}^\iagent[\action] - \bm{\reward}^\iagent[\action^*] = 0.91$. 
As \cref{fig:viz_ag1_no_interv} shows, this implies that the unintervened agent would almost never select $\action^*$ even when it has a higher exploration coefficient $\beta$.
Even with intervention policy S1, over $\horizon=200$, the \planner{} wouldn't be able to align the \agent{}'s preferred action with its own as shown in \cref{fig:viz_ag1_interv10}.
In contrast, \cref{fig:viz_ag1_intervT20} shows that a \planner{} with intervention policy S2 would outperform S1 and achieve a higher score, but the \agent{} eventually discovers its own preference when the intervention stops in S2 and then it no longer selects $\action^*$. 
This further highlights the extent of non-stationarity in the environment that affects the intervention policy of the \planner{} and also the learning behavior of the \agent{}. 
It also demonstrates the difficulty of the learning problem in our setup and the importance of learning a cost-efficient few-shot adaptable \planner{} intervention policy to effectively intervene on unknown adaptive agents.

\subsection{Description of baselines}%
\label{subsec:bandit_baselines}

We now describe the details of our evaluated baselines in \cref{sec:bandit-setting} along with their variations that assume access to an agent state oracle.

\paragraph{Rule based intervention with an \agent{} state oracle (RB):} 
Given an oracle that correctly identifies the action $\action_{\itime}$ to be taken by an \agent{} in the next time step, a simple rule based approach is for the \planner{} to intervene at time $\itime$ when $\action_{\itime} \not= \action^*$.
We assume that the \planner{} always intervenes with a fixed incentive ($\intervention = 0.5$ or $1$) and we compute the \planner{}'s maximum possible score. Note that this is not a realistic solution for the \planner{} since it is impractical to expect the availability of such an oracle, especially for out of distribution test agents.

\paragraph{Model-free learning based intervention policy:}
In this framework, we assume that the planner has a recurrent intervention policy that outputs a distribution over interventions $\action^\iplanner_\itime\sim\policy^\iplanner_\ppolicypar\brck{\action^\iplanner_\itime|\action^\iagent_{\itime-1}, \action^\iplanner_{\itime-1}, \hidden^\iplanner_{\itime-1}}$, conditioned on the planner's intervention and observed agent action at $\itime-1$. 
The policy network is trained using REINFORCE for the MF-RL baseline and using MAML for the MF-MAML baseline. 

\paragraph{Learned intervention policy with an \agent{} state oracle:}
In this setting, the \planner{} learns a recurrent intervention policy that outputs a distribution over interventions 
$
\action^\iplanner_\itime \sim \policy^\iplanner_\ppolicypar\brck{\action^\iplanner_\itime|\action^\iagent_{\itime}, \hidden^\iplanner_{\itime-1}}
$
conditioned on the true agent action at time $t$ provided by an oracle. 
The policy network is trained using REINFORCE for the SB-RL baseline and MAML for the SB-MAML baseline.

\paragraph{Learned intervention policy with a world model without meta-learning (WM-RL):}
In this setting, we use our proposed recurrent world model with a recurrent intervention policy trained using REINFORCE. 
Here, the policy network outputs a distribution over interventions
$
\action^\iplanner_\itime\sim\policy^\iplanner_\ppolicypar\brck{\action^\iplanner_\itime|\hat{\action}^\iagent_{\itime}, \hidden^\iplanner_{\itime-1}} 
$
where $\hat{\action}^\iagent_{\itime} = \argmax_\action \hat{\policy}_\apolicypar\brck{\action^\iagent_\itime|\action^\iagent_{\itime-1}, \action^\iplanner_{\itime-1}, \hidden^\iagent_{\itime-1}}$.

We would like to highlight an implementation detail in our baselines indicated `RL' in \cref{sec:bandit-setting}.
Since we evaluate our learnt \planner{} policy in the $\testK$-shot adaptation setting which is common in the meta-learning literature, we ensure that the \planner{} policies that are not meta-trained are also allowed to $\testK$-shot adapt at test time.
This means that the `RL' policies are also updated at test time, before evaluation, using $\testK$ rounds of \planner{}-agent interactions. 
This is in contrast to \cref{sec:single-shot-setting} where `RL' was trained from scratch during test time adaptation.
It further shows that even with pre-training (on the same set of train agents as used by `MAML'), standard policy gradient update does not lead to effective test time $\testK$-shot adaptation on test agents. 

\begin{table}[t]
\caption{
    \textbf{\capsPlanner{} (with oracle agent state input) scores across 3 random seeds.}
    These baselines are not applicable in practice since they cheat by assuming access to an oracle that always informs them of the agent's next action. 
    We include them here as a form of standardization with respect to a (perfect) system that does not face the challenges of partial observability or out-of-distribution generalization in our setting.
    }
    \centering
    \small
    \begin{tabular}{lrrrrr}%
    \textbf{Train on UCB, $\beta=0.17$} & Test on $\beta=0.17$ & $\beta=0.27$ & $\beta=0.42$ & $\beta=0.5$ & $\beta=0.67$ \\
    \emph{No intervention} & 3 (0) & 5 (0) & 8 (0) & 10 (0) & 12 (0) \\
    RB & 173 (0) & 166 (0) & 154 (0) & 146 (0) & 126 (0)\\
    SB-RL & 168 (3) & 138 (27) & 128 (26) & 122 (24) & 107 (22)\\
    SB-MAML & 169 (3) & 169 (1) & 155 (2) & 148 (1) & 128 (2)\\
    \midrule
    \textbf{Train on $\epsilon$-greedy, $\epsilon=0.1$} & $\epsilon=0.1$ & $\epsilon=0.2$ & $\epsilon=0.3$ & $\epsilon=0.4$ & $\epsilon=0.5$\\
    \emph{No intervention} & 3 (0) & 4 (1) & 7 (0) & 9 (1) & 11 (0) \\
    RB & 156 (3) & 130 (1) & 105 (4) & 87 (4) & 62 (6)\\
    SB-RL & 148 (2) & 119 (3) & 87 (4) & 75 (6) & 50 (2)\\
    SB-MAML & 152 (1) & 126 (2) & 105 (3) & 66 (3) & 30 (9) \\
    \midrule
    \textbf{Train on UCB, $\beta=0.67$} & $\beta=0.17$ & $\beta=0.27$ & $\beta=0.42$ & $\beta=0.5$ & $\beta=0.67$\\
    \emph{No intervention} & 3 (0) & 5 (0) & 8 (0) & 10 (0) & 12 (0)\\
    RB & 173 (0) & 166 (0) & 154 (0) & 146 (0) & 126 (0)\\
    SB-RL & 166 (3) & 163 (2) & 150 (3) & 146 (2) & 128 (2)\\
    SB-MAML & 173 (1) & 170 (0) & 159 (0) & 152 (0) & 133 (0) \\
    \midrule
    \textbf{Train on $\epsilon$-greedy, $\epsilon=0.5$} & $\epsilon=0.1$ & $\epsilon=0.2$ & $\epsilon=0.3$ & $\epsilon=0.4$ & $\epsilon=0.5$\\
    \emph{No intervention} & 3 (0) & 4 (1) & 7 (0) & 9 (1) & 11 (0) \\
    RB & 156 (3) & 130 (1) & 105 (4) & 87 (4) & 62 (6)\\
    SB-RL & 49 (46) & 51 (35) & 64 (29) & 61 (15) & 28 (17)\\
    SB-MAML & 93 (45) & 62 (32) & 32 (13) & 58 (25) & 24 (17) \\
    \end{tabular}
    \label{tab:oracle_results_table}
\end{table}

In \cref{tab:oracle_results_table}, we compare the test time scores for the \planner{} policy having access to a state based oracle.
We observe that overall, the meta-trained \planner{} policy (SB-MAML) achieves a higher score even with distribution shift across different bandit algorithms and different levels of exploration, compared to the SB-RL baseline. 
The rule based baseline also shows strong performance but we note its scores do not reflect adaptation to distribution shift.
However, none of these baselines that assume the \planner{} has access to an oracle that correctly predicts the agent's action at the next time step are realistic.
We can only treat the scores in \cref{tab:oracle_results_table} as gold standards in a perfect system that does not account for the challenges faced by a \planner{} in practice. 

\paragraph{Training details.}
In \cref{sec:single-shot-setting}, the \planner{} policy $\policy^\iplanner$ is a fully connected neural network (MLP) with one hidden layer and ReLU activation.
Given an (noisy) observed value of the agent type as input, it predicts the probability of intervention: $\policy^\iplanner_\itime$.
The \planner{}'s action at time $\itime$ is $\action^\iplanner_\itime \sim \Bernoulli{}\brck{\policy^\iplanner_\itime}$.

For the `RL' \planner{}, it is trained on the test agents starting from scratch over $\testK$ episodes before evaluation. 
For the MAML \planner{}, it is meta-trained to learn an initial parameterization with a different set of training agents and evaluated with $\testK$-shot adaptation on the test agents.

In \cref{sec:bandit-setting}, the recurrent world model and policy networks are GRUs with 2 layers and hidden state dimension 128. For meta-training, the inner gradient update loop uses SGD optimizer with a learning rate of $7\times10^{-4}$ whereas the meta-update step uses Adam with a learning rate of 0.001. The world model is trained only with the set of training agents, it is not adapted at test time: only the policy network is $\testK$-shot adapted. 

We plan to release the code for our implementation with the published paper.

\subsection{Overview of $\testK$-shot adaptation with \oursolution{}:}
\label{subsec:meta-test-algo}

\cref{algo:K-ADAPT} outlines our framework for $\testK$-shot adaptation of the meta-trained \planner{} to test agents. In our experiments, $\testK=1$. 

\begin{algorithm}[t]
\caption{\oursolution{} ($\testK$-shot Adaptation)}
\begin{algorithmic}[1]
\STATE{Initialize \planner{} with trained parameters ($\ppolicypar_{\textrm{meta}}$, $\apolicypar_{\textrm{train}}$), and hidden states $\hidden^\iagent_0, \hidden^\iplanner_0$. }

\FOR{agents (tasks) $\iagent = 1,\dots,\testsetsize$}
\STATE{Initialize \agent{}: ($\mean^\iagent,\policy^\iagent_0$), task specific intervention policy parameter  $\ppolicypar\brck{\task^\iagent_0}=\ppolicypar_{\textrm{meta}}$.
}
\FOR{$\optstep = 1,\dots,\testK$} 
\FOR{time \itime{} = $1,\dots,\horizon$}
\STATE{Predict $\hat{\action}^\iagent_{\itime} = \argmax_{\action^\iagent_\itime} \hat{\policy}_{\apolicypar_{\textrm{train}}}\brck{\action^\iagent_\itime | \action^\iagent_{\itime-1},\action^\iplanner_{\itime-1},\hidden^\iagent_{\itime-1}}$ using the world model.}
\STATE{Intervention: 
$\meanadj^\iagent = \mean^\iagent + \action^\iplanner_\itime, \quad
\action^\iplanner_\itime \sim \policy^\iplanner_{\ppolicypar\brck{\task^\iagent_\optstep}}\brck{ \action^\iplanner_\itime | \action^\iagent_{\itime-1}, \action^\iplanner_{\itime-1}, \hat{\action}^\iagent_\itime,\hidden^\iplanner_{\itime-1}}$.
}
\STATE{Agent acts: $\action^\iagent_\itime\sim\policy^\iagent_\itime$ and receives reward $\reward^\iagent_\itime \sim \normal\brck{\meanadj^\iagent, \sigma^2}$. $\policy^\iagent_{\itime} \mapsto \policy^\iagent_{\itime+1}$.} 
\ENDFOR{}
\STATE{Locally update $\ppolicypar\brck{\task^\iagent_\optstep} \mapsto \ppolicypar\brck{\task^\iagent_{\optstep+1}}$. \COMMENT{Using REINFORCE.} }\label{line:local_update_our_algo}%
\ENDFOR{}
\FOR{$\itime = 1,\dots,\horizon$}
\STATE{Predict $\hat{\action}^\iagent_{\itime} = \argmax_{\action^\iagent_\itime} \hat{\policy}_{\apolicypar_{\textrm{train}}}\brck{\action^\iagent_\itime | \action^\iagent_{\itime-1},\action^\iplanner_{\itime-1},\hidden^\iagent_{\itime-1}}$ using the world model.}
\STATE{Intervention: $\meanadj^\iagent = \mean^\iagent + \action^\iplanner_\itime, \quad
\action^\iplanner_\itime \sim \policy^\iplanner_{\ppolicypar\brck{\task^\iagent_{\testK}}}\brck{ \action^\iplanner_\itime | \action^\iagent_{\itime-1}, \action^\iplanner_{\itime-1}, \hat{\action}^\iagent_{\itime}, \hidden^\iplanner_{\itime-1}}$.
}
\STATE{Agent acts: $\action^\iagent_\itime\sim\policy^\iagent_\itime$, receives reward $\reward^\iagent_\itime \sim \normal\brck{\meanadj^\iagent, \sigma^2}$. Updates $\policy^\iagent_{\itime} \mapsto \policy^\iagent_{\itime+1}$.} 
\STATE{Update \planner{}'s score.}
\ENDFOR{}
\ENDFOR{}
\end{algorithmic}
\label{algo:K-ADAPT}
\end{algorithm}

\section{Additional experimental results with bandit agents}
\label{app:seeds}

\paragraph{Comparing the stability of \oursolution{} for different sets of random seeds.}
We trained and evaluated \oursolution{} for 3 additional random seeds in the $\testK = 1$-shot setting from \cref{tab:results_table}. 
In \cref{tab:additional_seeds_table}, we show the scores (mean and standard error) for both sets of seeds and all the six seeds combined. 
Our results indicate that the training and evaluation of \oursolution{} is stable, with reasonable and explainable variability across random seeds.
More specifically, models trained with different sets of random seeds result in similar mean scores with small standard error when $K=1$-shot evaluated on the less stochastic UCB agents.
In contrast, evaluation with the more stochastic $\epsilon$-greedy agents results in comparably larger standard error and more variation in the mean scores for models trained with different sets of random seeds.
Due to the computational costs involved, we were unable to evaluate all baselines with six seeds but on the basis of these observations, we do not expect a significant deviation from the claims made in \cref{sec:bandit-setting} and the values originally reported in \cref{tab:results_table} and \cref{tab:zero_shot_cross_table}, even with more seeds.

\begin{table}[t]
\caption{\textbf{Comparing the variability of \oursolution{} test scores with a trained model across different sets of random seeds.} Set 1 uses the seeds \{11, 26, 90\} and Set 2 uses the seeds \{12, 27, 91\}. `Combined' indicates mean and s.e. scores for models trained with the seeds \{11, 12, 26, 27, 90, 91\}. Overall, we observe that training \oursolution{} with different sets of random seeds shows little variance in the evaluation scores within error bounds, especially with the less stochastic UCB agent. For the $\epsilon$-greedy agent, we observe a higher standard error in the $K=1$-shot evaluation scores, but it is in line with the stochasticity associated with $\epsilon$-greedy action selection in the bandit agent.}
    \centering
    \begin{tabular}{lrrrrr}
         \textbf{Train on UCB, $\beta=0.17$} & Test on $\beta=0.17$ & $\beta=0.27$ & $\beta=0.42$ & $\beta=0.5$ & $\beta=0.67$ \\
         {\oursolution{} (Set 1; 3 seeds)} & 154 (2) & 151 (1) & 129 (1) & 129 (1) & 87 (0) \\
        {\oursolution{} (Set 2; 3 seeds)} & 144 (3) & 133 (3) & 122 (2) & 108 (2) & 81 (2) \\
         {\oursolution{} (Combined; 6 seeds)} & 151 (3)  & 142 (4) & 125 (2) & 118 (5) & 84 (2)\\
         \midrule
         \textbf{Train on UCB, $\beta=0.67$} & $\beta=0.17$ & $\beta=0.27$ & $\beta=0.42$ & $\beta=0.5$ & $\beta=0.67$\\
         {\oursolution{} (Set 1; 3 seeds)} & 132 (1) & 130 (1) & 123 (2) & 115 (2) & 99 (1)\\
           {\oursolution{} (Set 2; 3 seeds)}  & 119 (7) & 118 (5) & 110 (3) & 104 (3) & 87 (2) \\
           {\oursolution{} (Combined; 6 seeds)} & 126 (4) & 124 (3) & 116 (3) & 110 (3) &  93 (3) \\
        \midrule
        \textbf{Train on $\epsilon$-greedy, $\epsilon=0.1$} & $\epsilon=0.1$ & $\epsilon=0.2$ & $\epsilon=0.3$ & $\epsilon=0.4$ & $\epsilon=0.5$\\
         {\oursolution{} (Set 1; 3 seeds)} & 138 (1) & 112 (2) & 85 (3) & 66 (2) & 37 (4)\\
           {\oursolution{} (Set 2; 3 seeds)}  & 133 (2) & 133 (4) & 133 (4) & 133 (4) & 133 (4) \\
           {\oursolution{} (Combined; 6 seeds)} & 136 (2) & 123 (5) & 109 (11) & 99 (15) &  85 (22) \\
    \end{tabular}
    \label{tab:additional_seeds_table}
\end{table}

\end{document}